\def\paperID{0000}
\def\confName{CVPR}
\def\confYear{2024}

\def\paperTitle{\textbf{Multi-V2X: A Large Scale Multi-modal Multi-penetration-rate Dataset for Cooperative Perception}}

\def\authorBlock{
    Rongsong Li \qquad
    Xin Pei\thanks{Corresponding author} \qquad
    \\
    Tsinghua University \\
    {\tt\small lirs17@tsinghua.org.cn, peixin@tsinghua.edu.cn}
}

\newif\ifreview 
\newif\ifarxiv \newcommand{\arxiv}{\arxivtrue}
\newif\ifcamera 
\newif\ifrebuttal 

\arxiv %

\pdfoutput=1
\documentclass[10pt,twocolumn,letterpaper]{article}
\ifreview \usepackage[review]{cvpr} \fi
\ifarxiv \usepackage[pagenumbers]{cvpr} \fi
\ifrebuttal \usepackage[rebuttal]{cvpr} \fi
\ifcamera \usepackage{cvpr} \fi

\usepackage{graphicx}	
\usepackage{amsmath}	
\usepackage{amssymb}	
\usepackage{booktabs}
\usepackage{times}
\usepackage{microtype}
\usepackage{epsfig}
\usepackage[table,xcdraw,dvipsnames]{xcolor}
\usepackage{caption}
\usepackage{float}
\usepackage{placeins}
\usepackage{color, colortbl}
\usepackage{stfloats}
\usepackage{enumitem}
\usepackage{tabularx}
\usepackage{xstring}
\usepackage{multirow}
\usepackage{xspace}
\usepackage{url}
\usepackage{subcaption}
\usepackage{xcolor}
\usepackage[hang,flushmargin]{footmisc}
\usepackage{makecell}
\usepackage{algorithm}
\usepackage{algorithmic}

\ifcamera \usepackage[accsupp]{axessibility} \fi

\ifarxiv  \fi

\newcommand{\R}[1]{{%
    \textbf{%
        \ifstrequal{#1}{1}{\textcolor{red}{R#1}}{%
        \ifstrequal{#1}{2}{\textcolor{blue}{R#1}}{%
        \ifstrequal{#1}{3}{\textcolor{magenta}{R#1}}{%
        \ifstrequal{#1}{4}{\textcolor{teal}{R#1}}{%
                           \textcolor{cyan}{R#1}%
        }}}}%
    }%
}}

\usepackage{xr-hyper}

\makeatletter
\newcommand*{\addFileDependency}[1]{
  \typeout{(#1)}
  \@addtofilelist{#1}
  \IfFileExists{#1}{}{\typeout{No file #1.}}
}

\makeatother
\newcommand*{\myexternaldocument}[1]{
    \externaldocument{#1}
    \addFileDependency{#1.tex}
    \addFileDependency{#1.aux}
}

\definecolor{cvprblue}{rgb}{0.21,0.49,0.74}
\usepackage[pagebackref,breaklinks,colorlinks,citecolor=cvprblue]{hyperref}
\usepackage[capitalize]{cleveref}
\crefname{section}{Sec.}{Secs.}
\crefname{table}{Table}{Tables}
\crefname{figure}{Fig.}{Figs.}

\ifarxiv \crefname{appendix}{App.}{Apps.}
\else \crefname{appendix}{Suppl.}{Suppls.} \fi

\unless\ifarxiv \myexternaldocument{_supplementary} \fi

\begin{document}
\title{\paperTitle}
\author{\authorBlock}
\maketitle

\begin{abstract}

Cooperative perception through vehicle-to-everything (V2X) has garnered significant attention in recent years 
due to its potential to overcome occlusions and enhance long-distance perception. 
Great achievements have been made in both datasets and algorithms. 
However, existing real-world datasets are limited by the presence of few communicable agents, 
while synthetic datasets typically cover only vehicles. 
More importantly, the penetration rate of connected and autonomous vehicles (CAVs)
, a critical factor for the deployment of cooperative perception technologies, 
has not been adequately addressed.
To tackle these issues, we introduce Multi-V2X, a large-scale, multi-modal, 
multi-penetration-rate dataset for V2X perception. 
By co-simulating SUMO and CARLA, we equip a substantial number of cars and roadside units (RSUs) 
in simulated towns with sensor suites, and collect comprehensive sensing data. 
Datasets with specified CAV penetration rates can be obtained by masking some equipped cars as normal vehicles.
In total, our Multi-V2X dataset comprises 549k RGB frames, 146k LiDAR frames, 
and 4,219k annotated 3D bounding boxes across six categories. 
The highest possible CAV penetration rate reaches 86.21\%, with up to 31 agents in communication range, 
posing new challenges in selecting agents to collaborate with. 
We provide comprehensive benchmarks for cooperative 3D object detection tasks.
Our data and code are available at \textcolor{red}{https://github.com/RadetzkyLi/Multi-V2X}.
\end{abstract}

\section{Introduction}
\label{sec:intro}

Road traffic crashes lead to over 1.35 million deaths annually \cite{scanlon2021_waymo}, 
with 61.8\% attributable to human driver perception errors \cite{mueller2020_crash_ana}. 
To address this, researchers have focused on autonomous driving, 
particularly leveraging advancements in deep learning for perception tasks like object detection and semantic segmentation. 
Despite progress, individual perception struggles with occlusion and long-distance detection \cite{schoettle2017_sensor_fusion,liu2023_v2x_survey}. 
To guarantee safe, successful and effective driving task execution, cooperative or collabarative perception is proposed.
By integrating information from connected and autonomous vehicles (CAVs), roadside units (RSUs), etc., 
cooperative perception increases field of view (FOV), angle of view, and range of view greatly,
and thus obtains a more holistic understanding of the scene and brings benefits to downstream tasks.

Recent cooperative perception research has seen the development of simulation platforms \cite{xu2021_opencda}, 
synthetic \cite{xu2022_opv2v,xu2022_v2xvit,li2022_v2x-sim,wang2023_deepaccident} and 
real-world \cite{yu2022_dair_v2x,xu2023_v2v4real,hao2024_rcooper} datasets,
3D object deteection and tracking algorithms \cite{chen2019_cooper,chen2019_fcooper,wang2020_v2vnet,xu2022_v2xvit,
li2021_disconet,hu2022_where2comm,yang2024_how2comm,yang2023_what2comm,chiu_smith2023_selective_comm},
and adversial scenes \cite{xiang2023_v2xp_asg}, etc.

However, the following drawbacks of existing datasets impede further developments:
1) \textit{Limited agents in real-world datasets}. 
Real-world datasets often feature few interacting agents.
For example, one CAV and one RSU in DAIR-V2X \cite{yu2022_dair_v2x},
two CAVs in V2V4Real \cite{xu2023_v2v4real},
or only RSUs in RCooper \cite{hao2024_rcooper}. 
While useful for validation, they lack diversity for training purposes.
2) \textit{Limited categories in synthetic datasets}. 
Synthetic datasets predominantly include only cars 
(e.g., in the widely used OPV2V \cite{xu2022_opv2v}, V2XSet \cite{xu2022_v2xvit} and V2X-Sim \cite{li2022_v2x-sim}),
omitting vulnerable road users such as cyclists and pedestrians, 
which may lead cooperative perception algorithms
to deviate from autonomous driving's fundmental goal----zero crashes. 
3) \textit{Ignored CAV penetration rate}. 
To best know of the authors, there is no cooperative perception dataset noting this important concept.
In real-world datasets, there are only one or two CAVs, and several (randomly selected) in synthetic datasets.
If deployed, the number of CAVs is determined by CAV penetration rate, i.e., the ratio of CAVs in all motor vehicles running on the road.
The training data should be as close to the actual one as possible, so as to reduce domain gap.

We hence release the first multi-penetration-rate dataset Multi-V2X for cooperative perception to 
close in the real situation when deploying V2X. 
By co-simulation of CARLA \cite{dosovitskiy2017_carla} and SUMO \cite{lopez2018_sumo}, 
nearly all cars in the whole town are equipped with sensor suites.
All sensing data of equipped cars and RSUs in various towns are collected to form Multi-V2X,
containing 6 categories, 549k images, 146k point clouds, 4219k 3D bounding boxes.
By masking some equipped cars as normal vehicles, training datasets of specified CAV penetration rate 
(up to 86.21\% with a maximum connections of 31 in communication range) can be generated,
providing a realistic training ground for cooperative perception systems. 

Our contributions are summarized as follows:
\begin{itemize}
    \item Multi-V2X is the first multi-penetration-rate dataset, 
    supporting explorations of cooperative perception under various CAV penetration rates (up to 86.21\%). 
    And a masking algorithm is proposed to generate V2X dataset with specified CAV penetration rate.
    \item More than 549k images and 146k point clouds with 4219k annotated 3D boudning boxes 
    for 6 categories are provided in our Multi-V2X, enabling further exploitations.
    \item Comprehensive benchmarks for cooperative 3D object detection are reported.
\end{itemize}

\section{Multi-V2X Dataset}
\label{sec:dataset}

The \textbf{Multi-V2X} is a large-scale multi-modal, multi-penetration-rate, multi-categroy dataset.
We commence with the sensor suite, delineate the collection process, 
present data analysis, and detail the masking algorithm.

\subsection{Sensor suite on vehicles and RSUs}

\begin{figure}
    \centering
    \subcaptionbox{Sensors on cars \cite{xu2022_opv2v}}
      {\includegraphics[width=0.45\linewidth]{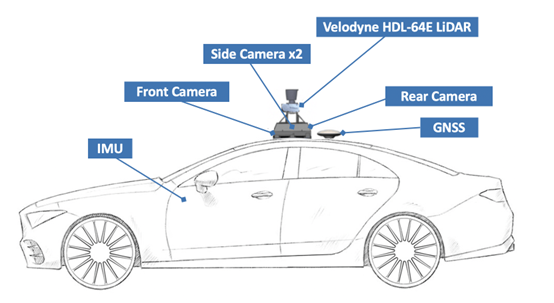}}
    \subcaptionbox{Sensors on RSUs}
      {\includegraphics[width=0.45\linewidth]{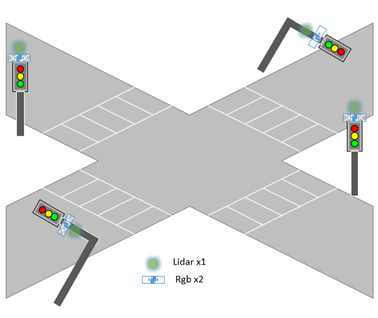}}
    \caption{Sensor layout in Multi-V2X}
    \label{fig:sensor-mounting-position}
\end{figure}

We target for autonomous cars (not trucks, buses, etc.) and hence only cars would be equipped
with sensor suite, i.e., 4 RGB cameras, 1 LiDAR, 1 sementic LiDAR, 1 GNSS. 
The installation way is exactly same as OPV2V \cite{xu2022_opv2v} and V2XSet \cite{xu2022_v2xvit},
that is, the four cameras (front, rear, left and right) and 2 LiDARs are on the roof of a car.
The traffic lights in intersections are regarded as RSUs and a random one is selected as RSU
if multiple traffic lights lie in a intersection. Each RSU is equipped with 2 RGB cameras,
1 LiDAR, 1 sementic LiDAR and 1 GNSS, away from the road surface 14 feets \cite{xu2022_v2xvit}.
The mounting angle of cameras of RSUs are manually determined to capture better road environment.
All sensors stream at 20Hz but record at 10Hz.
The illustrations of installation are depicted in Figure \ref{fig:sensor-mounting-position},
and configurations of sensors are in Table \ref{tab:dataset-sensor-details}.

\begin{table}[t]
  \centering
  \resizebox{\columnwidth}{!}{
    \begin{tabular}{ll}
      \toprule 
      \textbf{Sensor}   &  \textbf{Details}   \\
      \midrule 
      RGB camera    & FOV: 100$^{\circ}$, resolution: $800\times{600}$, frequency: 20Hz  \\
      \midrule
      \makecell{(Semantic) \\LiDAR}         & \makecell[l]{range: 120m, point: 1.3M points/s, horizental FOV: 360$^{\circ}$,\\ vertical FOV of car: 40$^{\circ}$(-30$^{\circ}$ $\sim$ 10$^{\circ}$), 
        \\vertical FOV of RSU: 40$^{\circ}$(-40$^{\circ}$ $\sim$ 0$^{\circ}$), \\rotation frequency: 20Hz} \\
      \midrule
      GNSS           & error: 0.02m  \\
      \bottomrule
    \end{tabular}
  }
  \caption{Sensor configurations in Multi-V2X}
  \vspace{-0.05in}
  \label{tab:dataset-sensor-details}
\end{table}

\subsection{CARLA-SUMO Co-simulation}

SUMO \cite{lopez2018_sumo} is a powerful micro traffic simulator and 
known for traffic flow simulation. 
CARLA \cite{dosovitskiy2017_carla} is an autonomous driving simumator and widely 
used to collect synthetic datasets for various perception tasks. 
Both are open-sourced.
To approximate the real sensing data and traffic movements, 
we leverage SUMO for traffic management (including route planning, 
action and signal control, etc.) and CARLA for sensor simulation and data recording.
The co-simulaton is progressed by synchronizing states of actors in SUMO to CARLA 
and then updating sensor simulation at each timestep.
For each town, vehicles and pedestrians are spawned and roam around in the town 
with randomly generated routes.
Hundreds of vehicles and pedestrians are spawned in six towns 
(Town01, Town03, Town06, Town07, Town10HD), covering crossroads,
T-junctions, segments, mid-blocks, rural roads, etc.
For each town, we manually selected 30s to 40s to record data,
which includes sensing data, annotations, movements of all actors during this period.

\subsection{Dataset Statistics}

As shown in Table \ref{tab:dataset-comparison}, 
Multi-V2X contains 549k images, 146k point clouds, 4219k 3D bounding boxes 
(lie in $x\in[-140, 140]$m, $y\in[-40, 40]$m of a agent \cite{xu2022_opv2v,xu2022_v2xvit}),
for 6 categories (car, van, truck, cyclist, motor, pedestrian),
supporting V2V and V2I cooperation. 
The maximum number of connections in 70m communication range \cite{xu2022_opv2v,xu2022_v2xvit} reaches 31, 
much greater than ever before, posing challenges to trade off performance and bandwidth. 
In fact, the number of connections reflects CAV penetration rate to some extent,
i.e., the former proportational to the latter. 
As estimated (see Appendix \ref{sec:appendix_cav}), CAV penetration rates of existing datasets are less than 20\%,
mostly in 10\% to 20\%, thus provide few insights for situation with high CAV penetration rate.
In contrast, Multi-V2X enables advance explorations for situation with up to 86.21\% CAV penetration rate.

\begin{figure*}
  \centering
  \begin{subfigure}{0.40\textwidth}
    \begin{minipage}[c]{\textwidth}
      \subcaptionbox{Connections=0}
        {\includegraphics[width=0.45\textwidth]{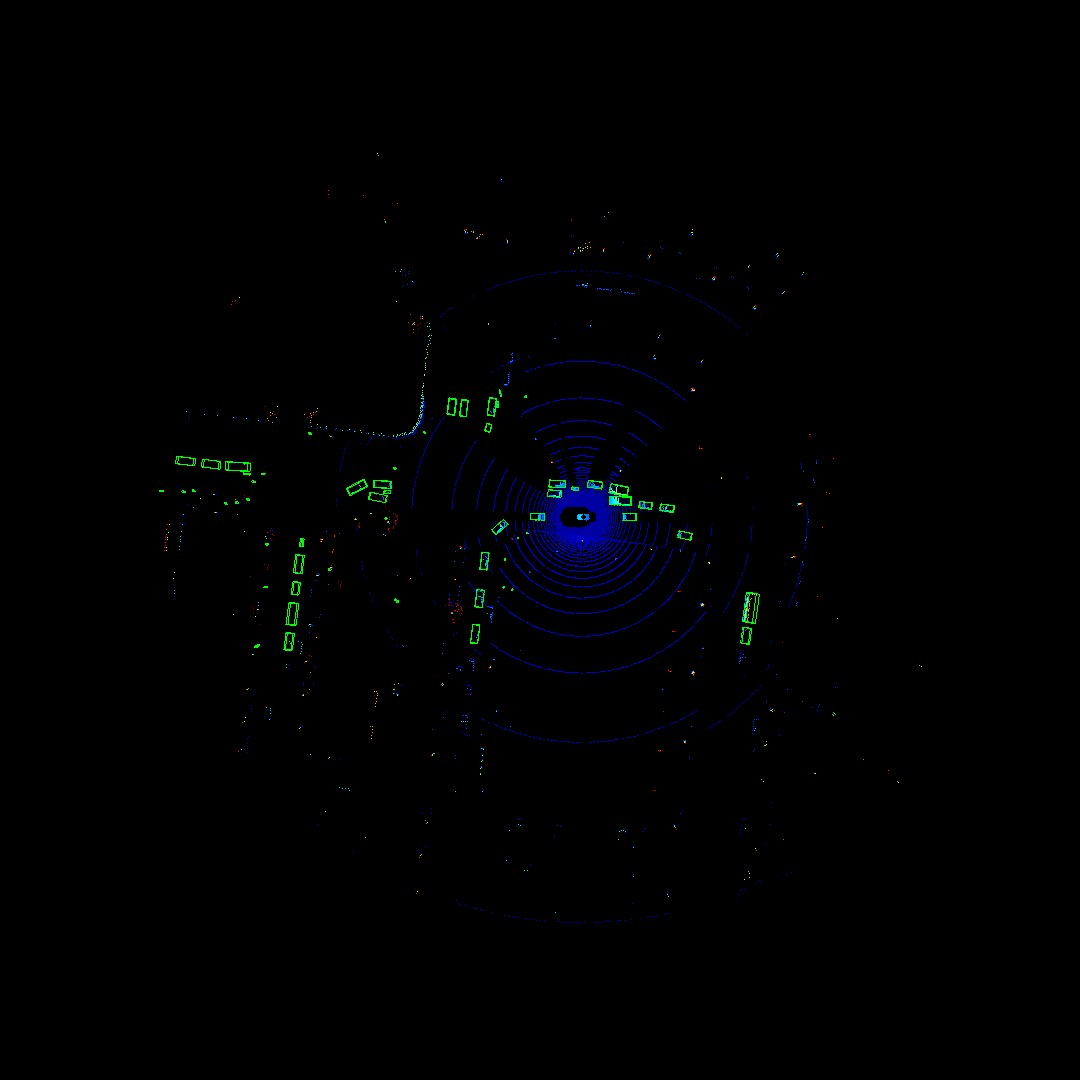}}
      \subcaptionbox{Connections=2}
        {\includegraphics[width=0.45\textwidth]{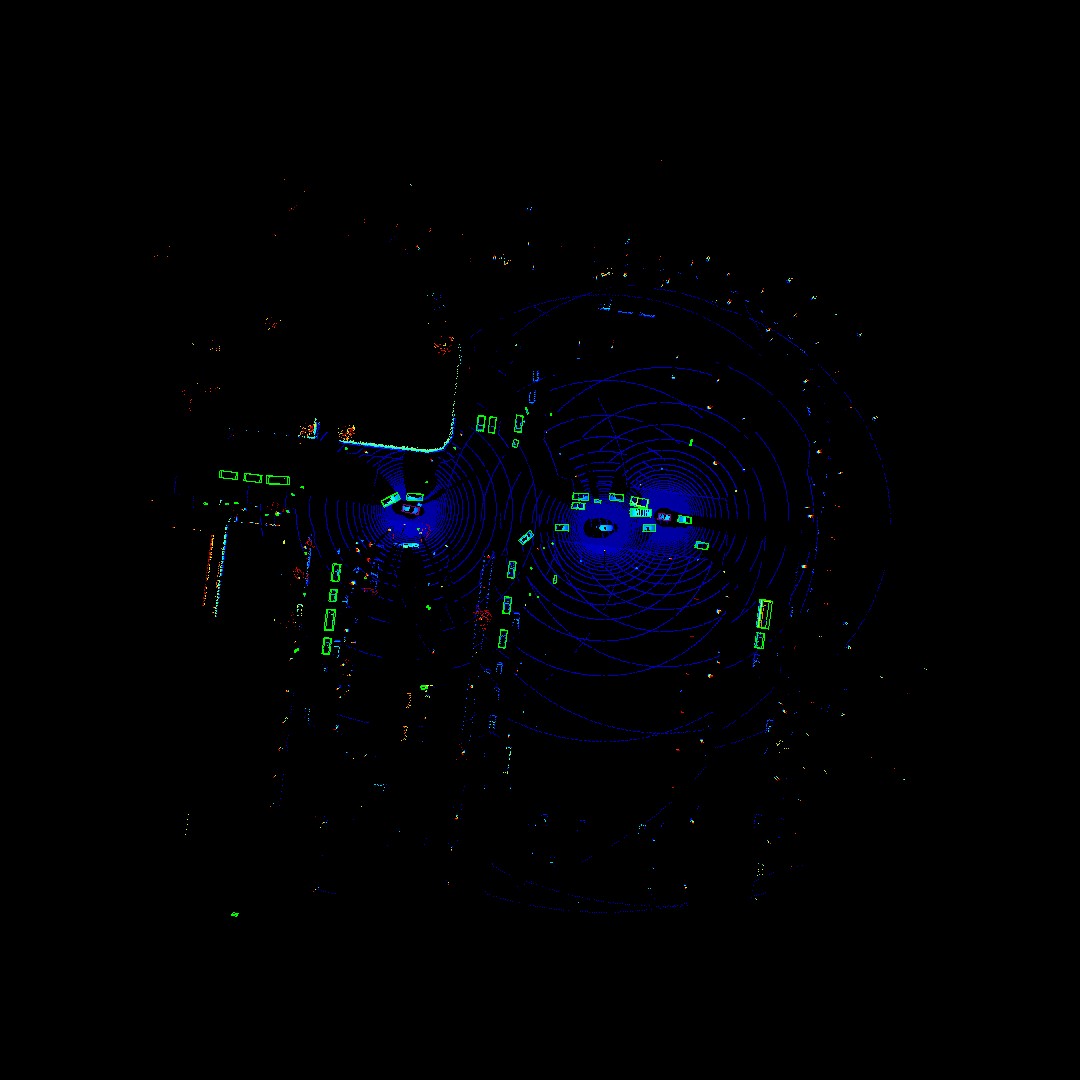}}
      \\
      \subcaptionbox{Connections=8}
        {\includegraphics[width=0.45\textwidth]{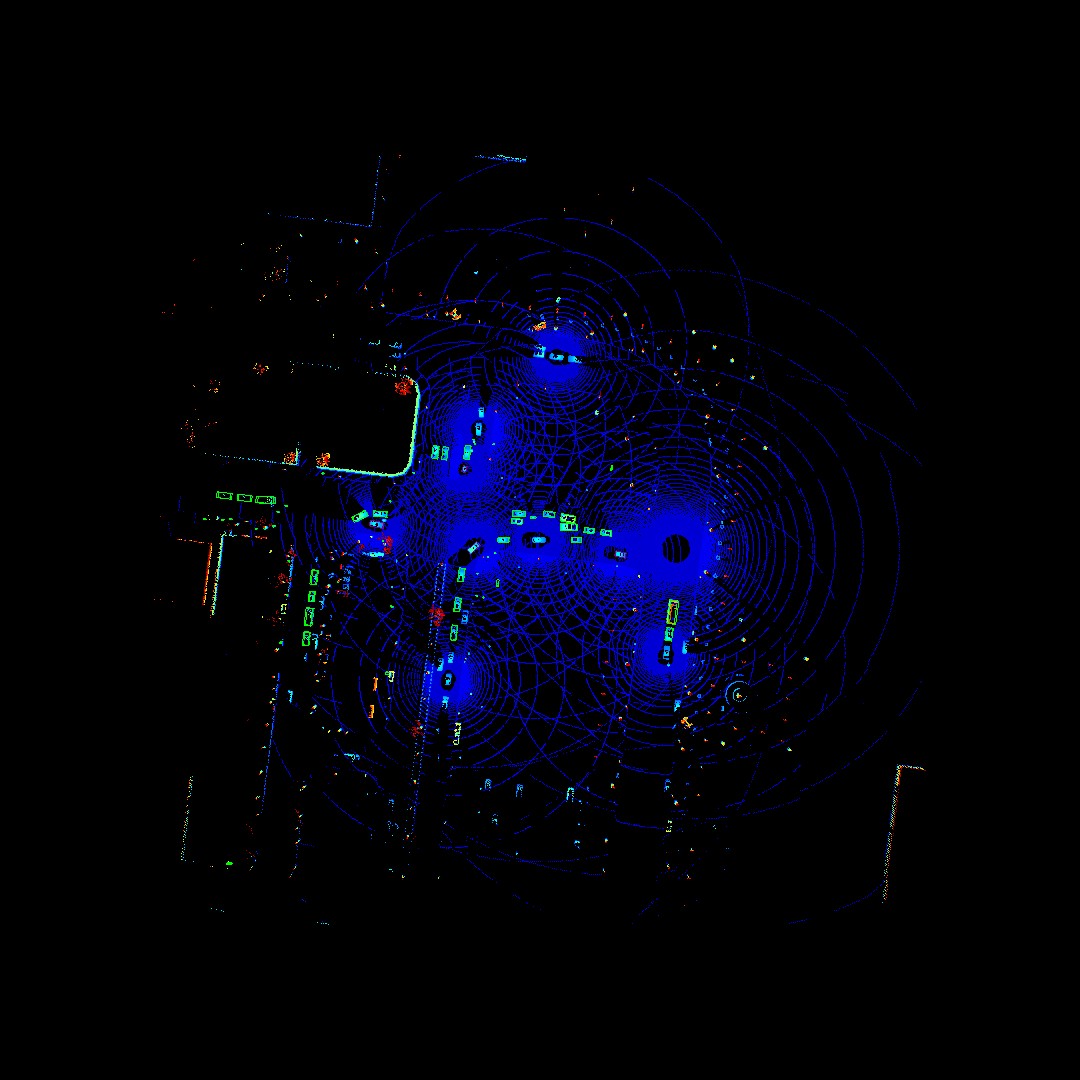}}
      \subcaptionbox{Connections=14}
        {\includegraphics[width=0.45\textwidth]{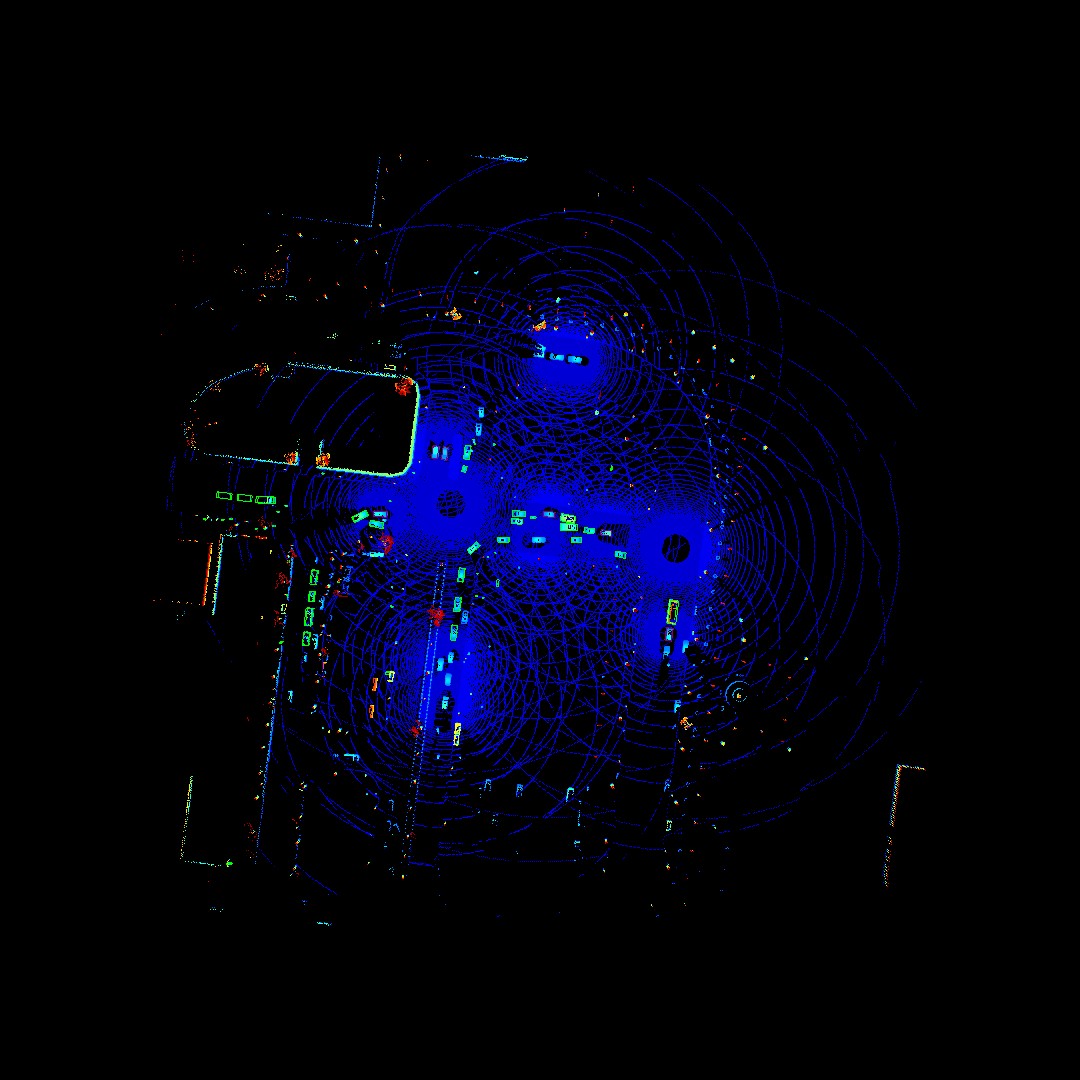}}
      \\
      \subcaptionbox{Connections=23}
        {\includegraphics[width=0.45\textwidth]{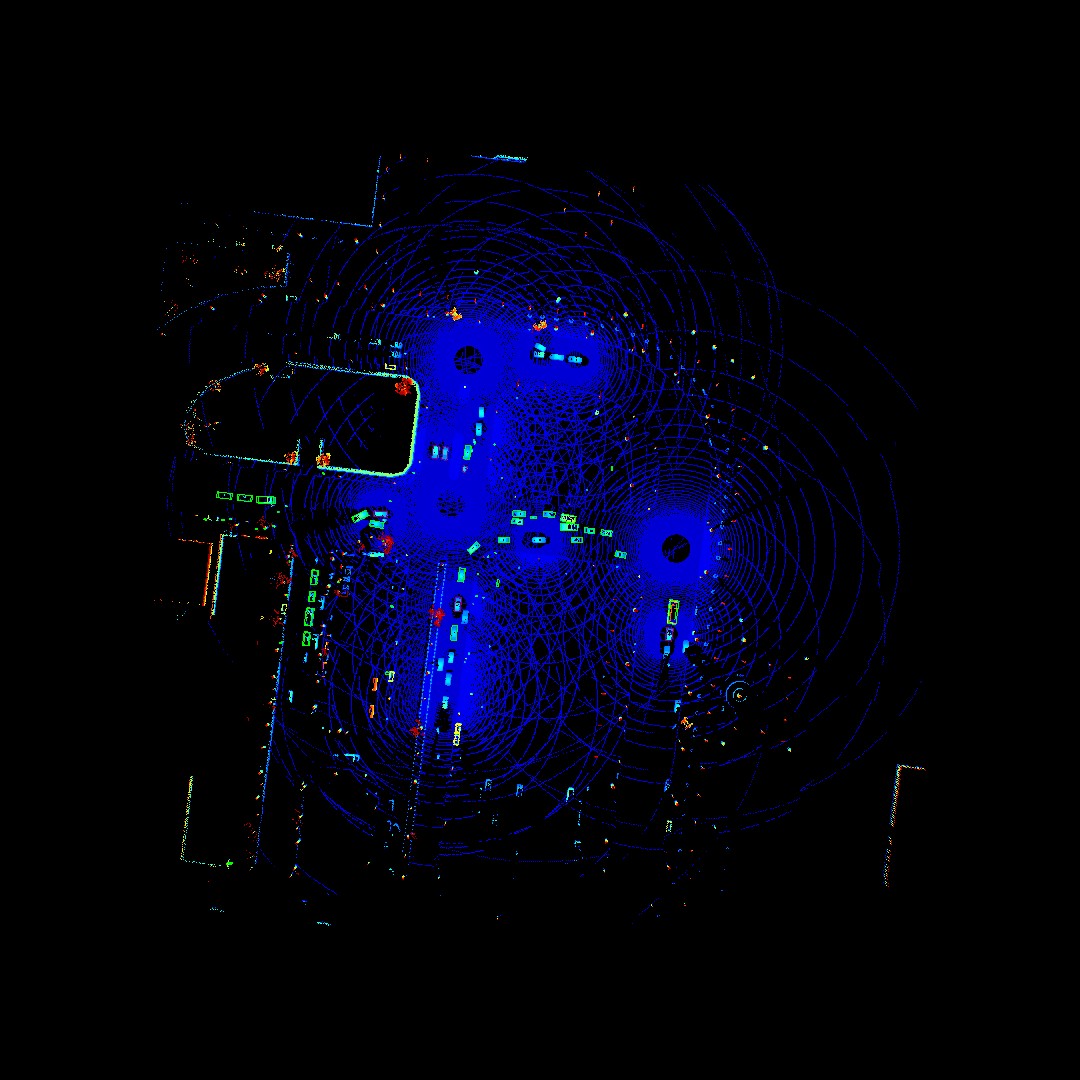}}
      \subcaptionbox{Connections=29}
        {\includegraphics[width=0.45\textwidth]{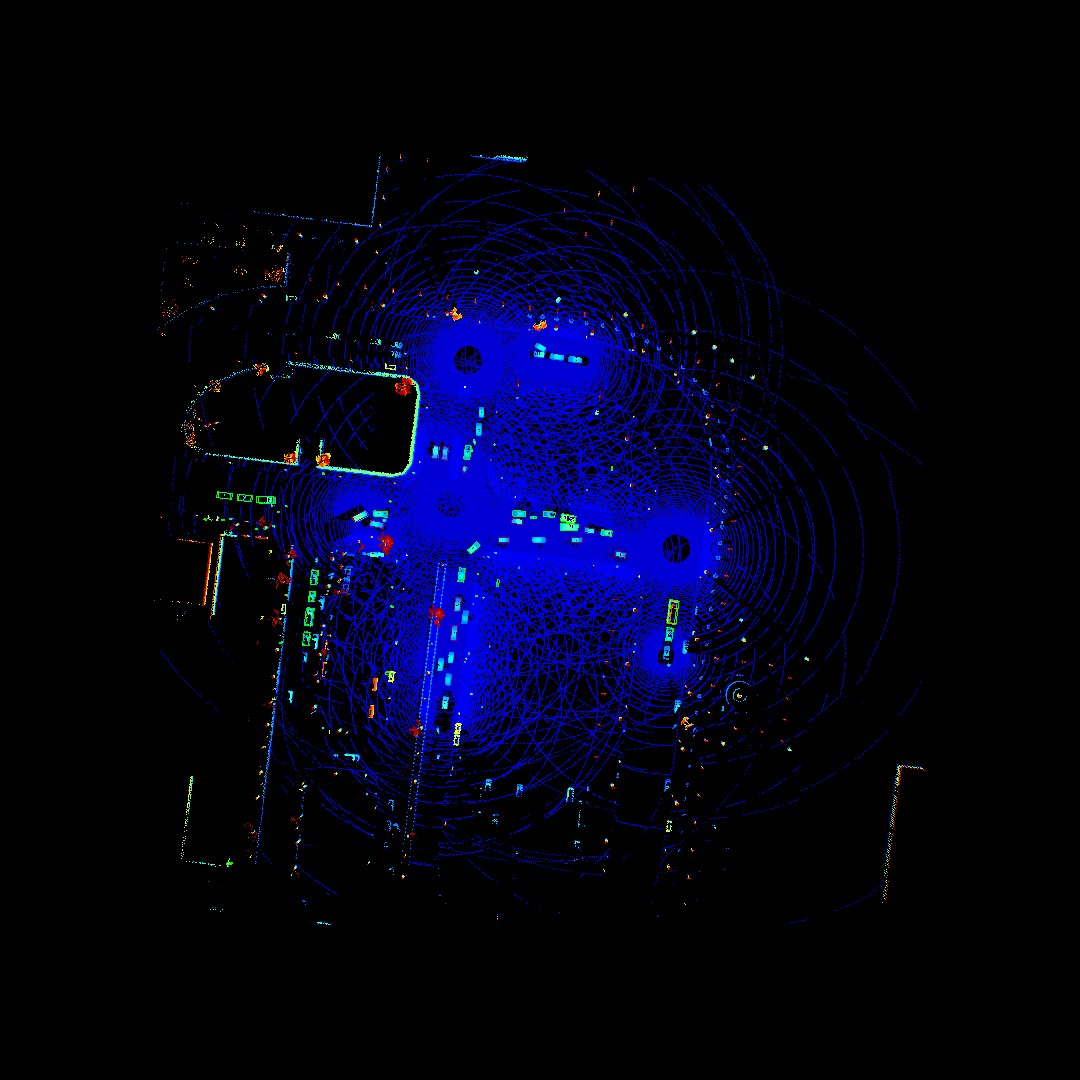}}
    \end{minipage}
  \end{subfigure}
  \begin{subfigure}{0.50\textwidth}
    \begin{minipage}[c]{\textwidth}
      \subcaptionbox{Category}
        {\includegraphics[width=0.45\textwidth]{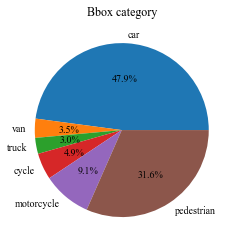}}
      \subcaptionbox{Count}
        {\includegraphics[width=0.45\textwidth]{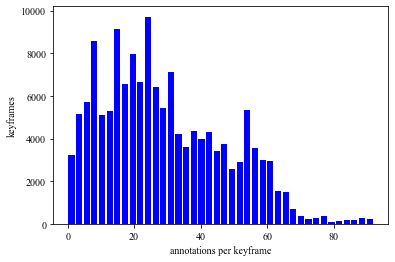}}
      \\      
      \subcaptionbox{Size}
        {\includegraphics[width=\textwidth]{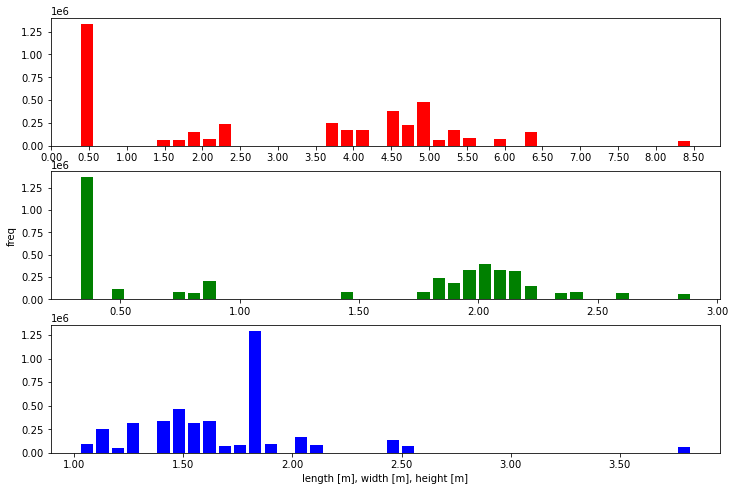}}
    \end{minipage}
  \end{subfigure}
  
  \caption{(a)-(f) Visualizations of bird's eye view point cloud when connections = 0, 2, 8, 14, 23 and 29 respectively.
    The more connections, the more clear CAV can sense the environment. (g) Statistics for bounding box categories. 
    (h) Counts for annotations per keyframe. (i) Statistics for bounding box sizes.}
  \label{fig:multiv2x-bbox-overview}
\end{figure*}

As shown in Figure \ref{fig:multiv2x-bbox-overview}, 
the more connections (implicating higher CAV penetration rate), the more clear the CAV can sense the environment, 
but the marginal effect decreases, which calls for new methods to select the best collaborators
or features while ensuring communication bandwidth.
The 3D bounding boxes cover a variety of categories, sizes, numbers, making it closer to real-world situations.
Minimum, maximum and average number of 3D bounding boxes per frame are 0, 92, 28.9 respectively.

\begin{table*}[t]
  \centering
  \resizebox{\linewidth}{!}{
    \begin{tabular}{llllllllll}
      \toprule
      \textbf{Dataset}                  & \textbf{Year} & \textbf{Source} & \textbf{V2X}    & \makecell[l]{\textbf{RGB Images}} & \textbf{LiDAR} & {\makecell[l]{\textbf{3D boxes}}} & \textbf{Categories} & \textbf{Locations}  & \textbf{Connections}   \\
      \midrule
      DAIR-V2X \cite{yu2022_dair_v2x}    & 2022 & real & V2I    & 71k        & 71k   & 1200k    & 10  & Beijing,China     & 1  \\
      V2V4Real \cite{xu2023_v2v4real}    & 2023 & real & V2V    & 40k        & 20k   & 240k     & 5   & Ohio,USA         & 1    \\
      RCooper \cite{hao2024_rcooper}     & 2024 & real & I2I    & 50k        & 30k   & -        & 10  & -           &  -       \\
      OPV2V \cite{xu2022_opv2v}          & 2022 & sim  & V2V    & 132k       & 33k   & 230k     & 1   & CARLA       & 1-6     \\
      V2XSet \cite{xu2022_v2xvit}        & 2022 & sim  & V2V\&I & 132k       & 33k   & 230k     & 1   & CARLA       & 1-4    \\
      V2X-Sim \cite{li2022_v2x-sim}      & 2022 & sim  & V2V\&I & 283k       & 47k   & 26.6k    & 1   & CARLA       & 1-4   \\
      \midrule
      \textbf{Multi-V2X}                & \textbf{2024} & \textbf{sim} & \textbf{V2V\&I} & \textbf{549k}   & \textbf{146k}  & \textbf{4219k}  & \textbf{6}   & \textbf{CARLA}       & \textbf{0-31}  \\
      \bottomrule
    \end{tabular}
  }
  \caption{Comparisons among the representative public cooperative perception datasets.}
  \label{tab:dataset-comparison}
\end{table*}

\subsection{CAV Penetration Rate}

The maximum CAV penetration rates over the six towns range from 55.17\% to 86.21\%,
providing fundations for further explorations.
To reduce data redundancy, instead of running co-simulation and collecting data
for different CAV penetration rates, we propose to achieve various CAV penetration rate 
in one dataset. 
Specially, when co-simulation, many cars are equipped with sensors and then collect data normally.
When using, some equipped cars are masked as non-equipped cars,
so as to obtain dataset of specified CAV penetration rate.
The maximum CAV penetration rate is reached if no equipped car is masked.

Algorithm \ref{alg:construct-training-dataset} describe this process in detail. 
By Algorithm \ref{alg:construct-training-dataset}, dataset with target CAV penetration rate 
$r$ can be constructed from Multi-V2X. 
The threshold $r^{\text{zero}}_{\text{thr}}$ is used to ensure the selected car 
has connections with other agents most of the time, avoiding ineffective training.
When $r$ is sma1l, to obtain sufficient training samples, 
one can repeat Algorithm \ref{alg:construct-training-dataset} for many times with 
different random seeds. 
Taking Town10HD as an example, there are 58 motor vehilcles in total and 50 of them equipped with sensors. 
If $r=10\%$, then 5 equipped cars are selected as CAVs, 
and their data constitute a dataset with CAV penetration rate of 8.62\%(5/58) .

Considering that one equipped car may be selected for many times 
(if Algorithm \ref{alg:construct-training-dataset} is repeated), 
to avoid information leakage, we recommend to split training and test set along
time axis instead by scenarios like in OPV2V \cite{xu2022_opv2v} and V2XSet \cite{xu2022_v2xvit}.
For example, there are 30s data of an ego car, the first 20s as training part 
and the rest 10s as test part.
For efficient training, the ego car should be moving (e.g., speed greater than 2 m/s).

\begin{algorithm}
  \caption{Construction of dataset with specified CAV penetration rate}
  \label{alg:construct-training-dataset}
  \small
  \begin{algorithmic}
    \REQUIRE Target CAV penetration rate $r$ , Multi-V2X Dataset $\mathcal{D}^{\text{Multi-V2X}}$, threshold $r^{\text{zero}}_{\text{thr}}$
    \ENSURE The resulting dataset $\mathcal{D}^{\text{Multi-V2X}}_{r}$

    Initialize target dataset: $\mathcal{D}^{\text{Multi-V2X}}_{r} \gets \emptyset$

    \FOR{Data of each map $\mathcal{D}^{\text{map}}$ in $\mathcal{D}^{\text{Multi-V2X}}$}
      \STATE \textbf{Step 1: Stats gaining}
      \STATE \quad 1.1: Get the number of motor vehicles $N^{\text{veh}}$ and list of equipped cars $L^{\text{av}}$

      \STATE \textbf{Step 2: Node filtering}
      \STATE \quad 2.1 Count ratio of zero connections $r^{\text{zero}}$ for each node in $L^{\text{av}}$
      \STATE \quad 2.2 For ecah node, if $r^{\text{zero}} < r^{\text{zero}}_{\text{thr}}$, then add the node to candidate list $L^{\text{cand}}$

      \STATE \textbf{Step 3: Node selection/masking}
      \STATE \quad 3.1 Calculate expected number of CAVs: $N^{\text{cav}} = \lfloor N^{\text{veh}} \times r \rfloor$
      \STATE \quad 3.2 Randomly sample $N^{\text{cav}}$ nodes from $L^{\text{cand}}$ without replacement to be $L^{\text{cav}}$ 

      \STATE \textbf{Step 4: Data selection}
      \STATE \quad 4.1 Add data of RSUs: $\mathcal{D}^{\text{Multi-V2X}}_{r} \gets \mathcal{D}^{\text{Multi-V2X}}_{r} \cup \mathcal{D}^{\text{map}}_{\text{rsu}}$
      \STATE \quad 4.2 Add data of CAVs:  $\mathcal{D}^{\text{Multi-V2X}}_{r} \gets \mathcal{D}^{\text{Multi-V2X}}_{r} \cup \mathcal{D}^{\text{map}}_{L^{\text{cav}}}$
    \ENDFOR
  \end{algorithmic}
\end{algorithm}

\section{Experiments}
\label{sec:exper}

\subsection{Task and Metrics}

\textbf{Task Description} \quad
Multi-V2X supports cooperative 3D object detection and tracking and we focus on the former in this paper. 
Cooperative 3D object detection task requires leveraing sensing data from multiple views from multiple agents to 
detect poses of objects in corresponding area. 
The pose can be denoted as $[x, y, z, l, w, h, \theta]$ 
where $x, y, z$ denote coordinate of the center of an object in ego's coordinate system, 
$l, w, h$ denote length, width and height of the object, 
and $\theta$ denotes heading angle or yaw of the object.

\vspace{\baselineskip}
\noindent \textbf{Metrics} \quad
The common metric average precision (AP) is used to measure algorithm's performance, 
taking recall and precision into account. 
AP lies [0,1] and coser to 1 means better. 
The evaluation area is the rectangle ($x\in[-140, 140]$m, $y\in[-70, 70]$) centered at  the ego.

\subsection{Benchmark Models}

Using PointPillars \cite{lang2019_pointpillars}  as backbone, 
state-of-the-art cooperative methods of the following four fusion strategies are considered.
\begin{itemize}
    \item No Fusion: There is no data sharing among agents and the ego relies itself to perform detection. 
    This is individual perception and used for baseline.
	\item Late Fusion: The agents share predictions, by which final outputs are produced by non-maximum suppression.
	\item Early Fusion:  The raw sensing data shared by various agents are projected into ego's space and 
    then processing pipeline of No Fusion is applied.
	\item Intermediate Fusion: Each agent processes its own sensing data to 
    obtain intermediate feature mapping, which afterward is shared to other agents. 
    Next, the agent fuses these mappings to generate final outputs. 
    Two representative intermediate fusion methods are adopted, 
    i.e., V2X-ViT \cite{xu2022_v2xvit} and Where2comm \cite{hu2022_where2comm}. 

\end{itemize}

\subsection{Experiment Details}

Due to lack of algorithms targeted for high CAV penetration rate, 
we just construct a dataset $\mathcal{D}_{10\%}^{\text{Multi-V2X}}$ with 10\% CAV penetration rate from Multi-V2X 
by running Algorithm \ref{alg:construct-training-dataset} for 7 times. 
The resulting $\mathcal{D}_{10\%}^{\text{Multi-V2X}}$ contains 14943 frames (counted by 48 ego cars),
and the training, validation and test set are splited by 6:2:2 along time axis, 
resulting 8962:2987:2994 
(similar to OPV2V \cite{xu2022_opv2v} and V2XSet \cite{xu2022_v2xvit}). 
An ego car may have 0 to 8 connections overtime.
Except for the added 5 anchors for extra 5 categories, 
training config is exactly same as V2X-ViT \cite{xu2022_v2xvit}.
That is, both training and evaluation are under perfect settings.
The communication range is 70m, 
and 4 random agents are selected to collaborate with if connections are bigger than 4.
The voxel resolution is 0.4m for both height and width in PointPillar backbone.
Adam \cite{kingma2014_adam} optimizer with an initial learning rate of 0.001 is adopted 
and the learning rate is decayed every 10 epochs by a factor of 0.1.

\subsection{Benchmark Analysis}

As can be seen from Table \ref{tab:ap-multiv2x-10percent} and \ref{tab:recall-multiv2x-10percent},
the AP is low, caused by low recall rate of cycle and pedestrian.
The underlying reason is that the listed algorithms only leverage point cloud for detection, 
which has been proved to be insufficient for small objects
and needs integration with images \cite{xu2018_pointfusion}.

\begin{table}
    \centering
    \begin{tabular}{lllll}
        \toprule
        \textbf{Method}    &  \textbf{AP@0.3}   & \textbf{AP@0.5}  & \textbf{AP@0.7}  \\
        \midrule 
        No Fusion          & 0.307              & 0.237            & 0.117   \\
        Late Fusion        & 0.346              & 0.270            & 0.141  \\
        Early Fusion       & 0.510              & 0.408            & 0.235    \\
        \midrule 
        V2X-ViT \cite{xu2022_v2xvit}  & 0.440    & 0.350            & 0.228   \\
        Where2comm \cite{hu2022_where2comm} & 0.452  & 0.348        & 0.213   \\
        \bottomrule
    \end{tabular}
    \caption{Cooperative 3D object detection benchmarks on $\mathcal{D}_{10\%}^{\text{Multi-V2X}}$ .}
    \label{tab:ap-multiv2x-10percent}
\end{table}
  
\begin{table}
    \centering
    \resizebox{\linewidth}{!}{
    \begin{tabular}{lllllllll}
      \toprule 
      \textbf{Method}    &  \multicolumn{7}{l}{\textbf{Recall} (IoU=0.5)} \\
      ~                  &  \textbf{Car}   & \textbf{Van}  & \textbf{Truck}  & \textbf{Motor} &\textbf{Cycle}  & \textbf{Pedestrian}  & \textbf{Overall} \\
      \midrule
      No Fusion             & 0.626  & 0.619 & 0.368 & 0.305 & 0.025 & 0.192 & 0.426 \\
      Late Fusion           & 0.738  & 0.692 & 0.445 & 0.368 & 0.053 & 0.235 & 0.496  \\
      Early Fusion          & 0.880  & 0.811 & 0.549 & 0.716 & 0.208 & 0.323 & 0.634 \\
      \midrule 
      V2X-ViT\cite{xu2022_v2xvit} & 0.709 & 0.392 & 0.418 & 0.522 & 0.118 & 0.173 & 0.464 \\
      \midrule 
      Where2comm\cite{hu2022_where2comm}  & 0.683 & 0.400 & 0.431 & 0.323 & 0.036 & 0.063 & 0.391 \\
      \bottomrule
    \end{tabular}
    }
    \caption{Per category recall on $\mathcal{D}_{10\%}^{\text{Multi-V2X}}$ .}
    \label{tab:recall-multiv2x-10percent}
\end{table}

\section{Conclusion}
\label{sec:conclusion}

The important CAV penetration rate is neglected, 
impeding further explorations especially for V2X deployment in real-world. 
To tackle this, we release Multi-V2X, the first larget-scale, multi-penetration-rate cooperative dataset, 
consisting of 549k images, 146k point clouds, 4219k 3D bounding boxes for 6 categories, 
with up to 86.21\% CAV penetration rate. 
Multi-V2X is expected to boost current cooperative peception researches
and provide possibilities for handling various CAV penetration rates. 
Competative benchmarks are provided to pave the way for subsequent works.

\vspace{\baselineskip}
\noindent \textbf{Future Work} \quad 
For future work, it's worth developing algorithms aiming to 
select key collaborators or features to collaborate with especially 
for high CAV penetration rate situations 
so as to balance performance and communication bandwidth. 
In addition, in the matter of safe autonomous driving, 
vulnerable road users such as penetrians deserve more attention.

{\small
\bibliographystyle{ieeenat_fullname}
\bibliography{11_references.bib}
}

\ifarxiv \clearpage \appendix 
\maketitlesupplementary

\section{CAV Penetration Rate}
\label{sec:appendix_cav}

Take data in Town10HD as an example, the connections varies with the CAV penetration rate.
From Table \ref{tab:conn-under-penetration}, it seems that a CAV can always collaborate with at least another agent 
once the CAV penetration rate reaches 50\%.
With highest CAV penetration rate, a CAV and RSU can connect with at most 30 and 31 other agents
in communication range (70m) respectively.
Inferring from Table \ref{tab:dataset-comparison} and Table \ref{tab:conn-under-penetration}, 
the CAV penetration rate of 
OPV2V, V2XSet and V2X-Sim is less than 20\%. 
Therefore, existing datasets cannot support researches on high penetration rate.

\begin{table}[t]
    \centering
    \begin{tabular}{llllllllll}
        \toprule
        ~  & \multicolumn{9}{l}{\textbf{CAV penetration rate}}  \\
        ~  & \textbf{0.086} & \textbf{0.190} & \textbf{0.293} & \textbf{0.379} & \textbf{0.500} 
        & \textbf{0.586} & \textbf{0.690}  & \textbf{0.793}  & \textbf{0.862}  \\
        \midrule 
        Min connections  & 0     &  0    & 0     & 0     & 1     &  1     &  1     & 1      & 1  \\
        Max connections  & 3     & 6     & 10    & 12    & 16    & 21     & 23     & 28     & 30  \\
        Avg connections & 2.622 & 4.175 & 5.619 & 6.902 & 8.735 & 10.703 & 12.956 & 15.259 & 16.319  \\
        \bottomrule
    \end{tabular}
    \caption{Connections of CAVs varies with CAV penetration rate.}
    \label{tab:conn-under-penetration}
\end{table}

\section{Dataset Visualizations}

Figure \ref{fig:appendix-multiv2x-examples} and Figure \ref{fig:appendix-multiv2x-examples-2} give more examples in bird's eye view for each town,
where a green box denotes an object. 
For the sake of clarity, point clouds from at most 4 agents would be drawed.
Intuitively, our Multi-V2X covers diverse roadway types and traffic situations,
laying the foundations of developing cooperative perception algorithms.

In CARLA, Town01 is a basic town composed of T-junctions.
Town03 is the most complex, with a 5-spot junction, a roundabout, a tunnel.
Town05 is a squared-grid town with cross junctions and bridges.
Town06 has highways. 
Town07 is a rual environment with narrow roads and hardly traffic lights.
Town10HD is a city environment with realistic textures.

\begin{figure*}
    \centering
    \begin{subfigure}{0.30\textwidth}
        \begin{minipage}[c]{\textwidth}
            \centering
            \includegraphics[width=\textwidth]{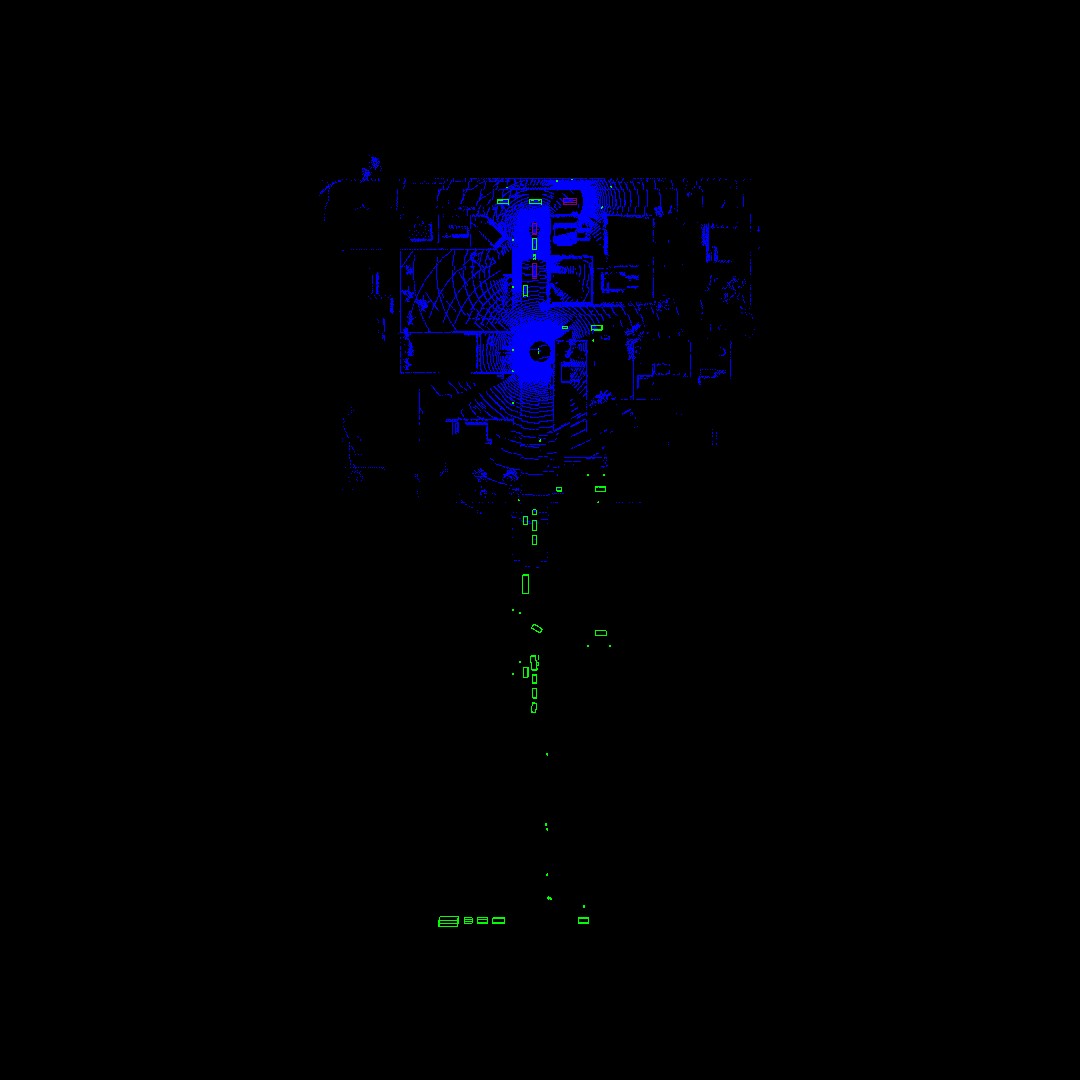} \\
            \includegraphics[width=\textwidth]{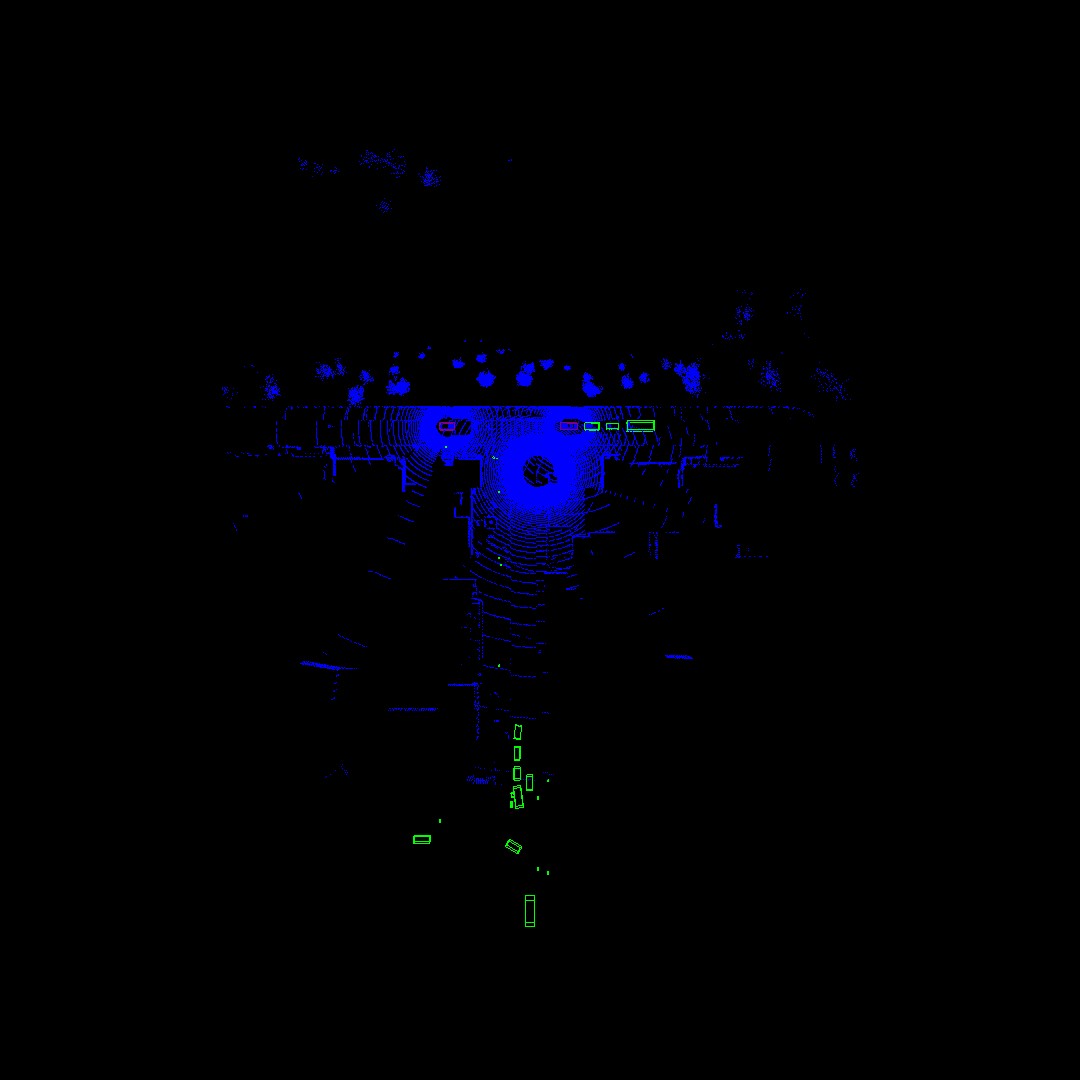} \\
            \includegraphics[width=\textwidth]{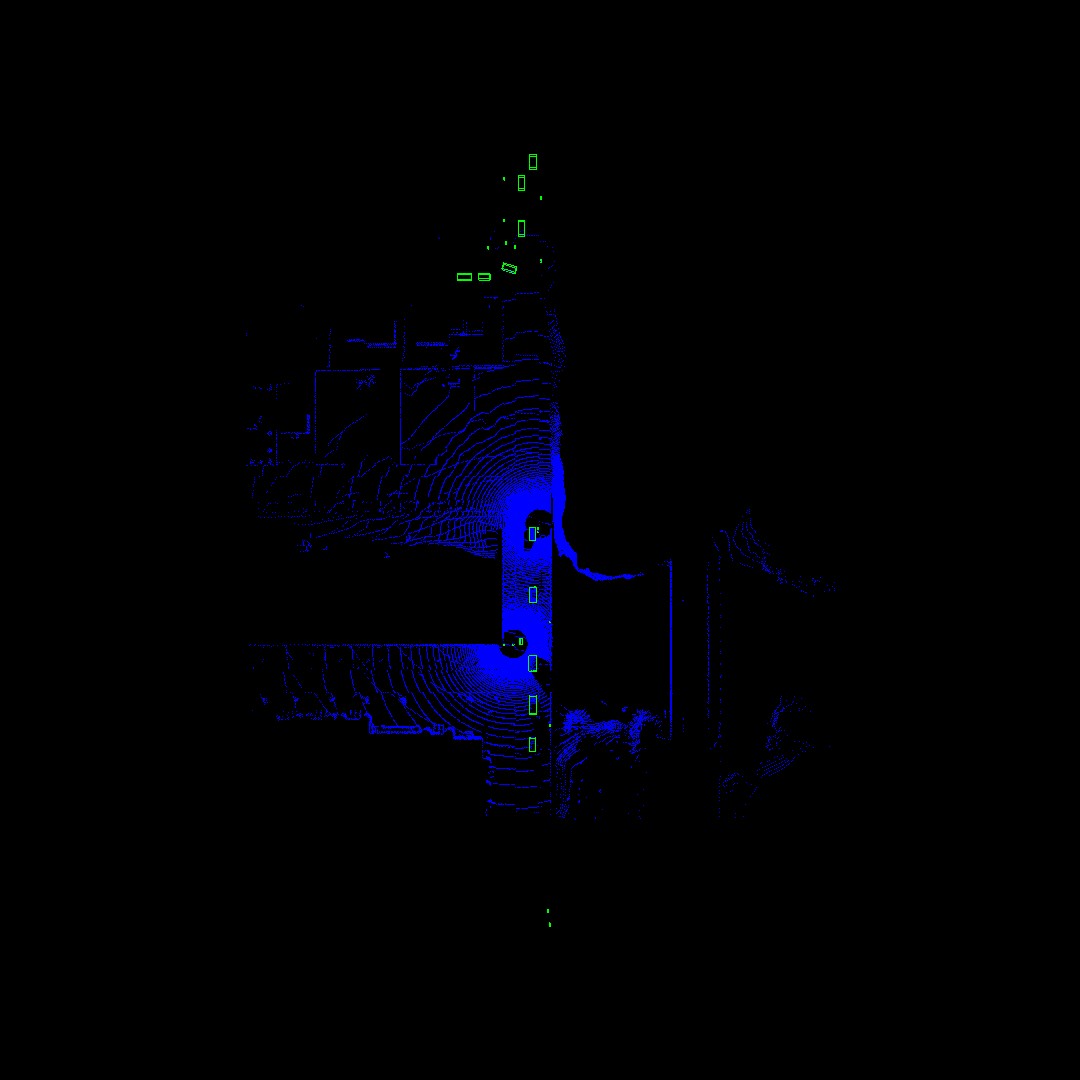} \\
            \includegraphics[width=\textwidth]{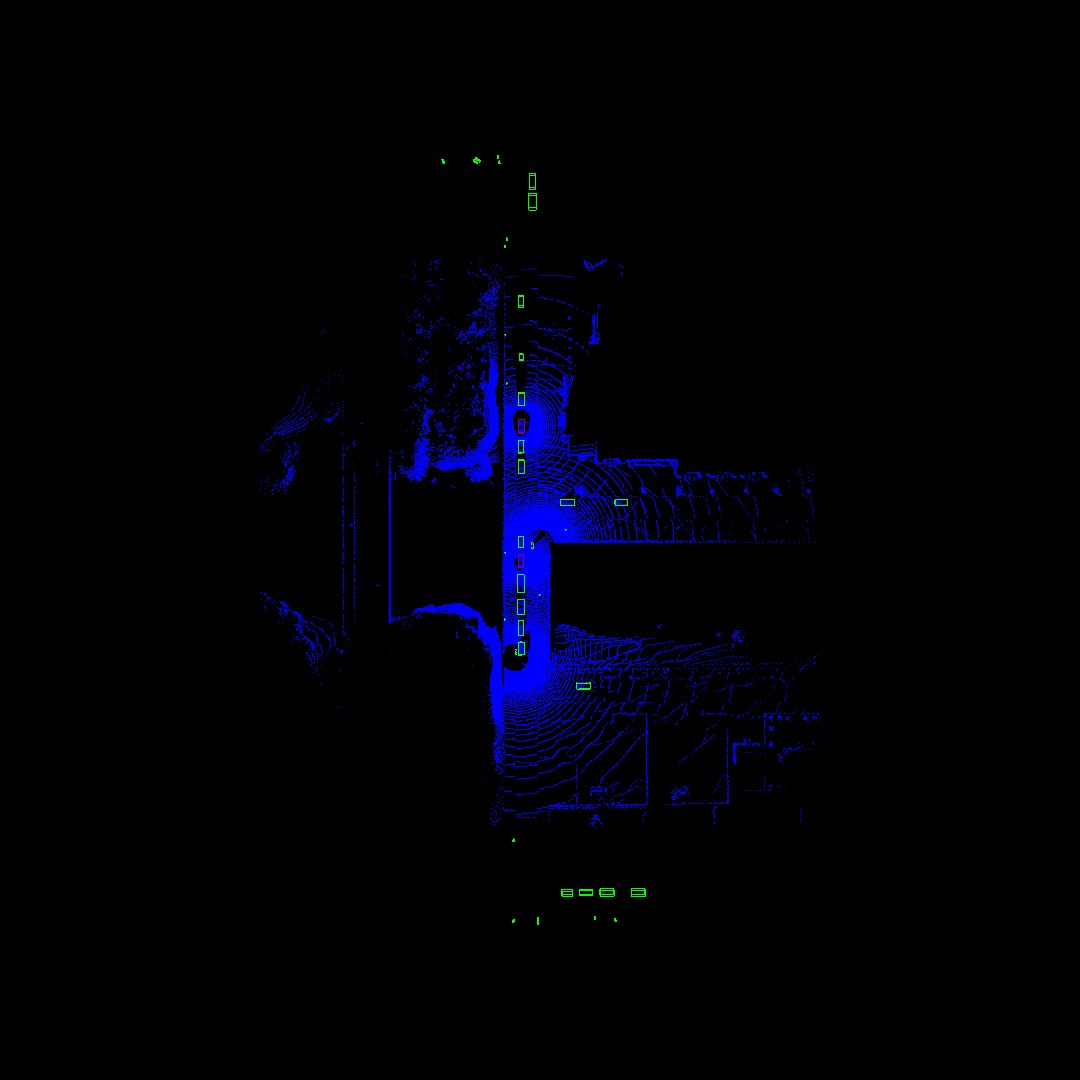} \\
            \caption{Town01}
        \end{minipage}
    \end{subfigure}
    \begin{subfigure}{0.30\textwidth}
        \begin{minipage}[c]{\textwidth}
            \centering
            \includegraphics[width=\textwidth]{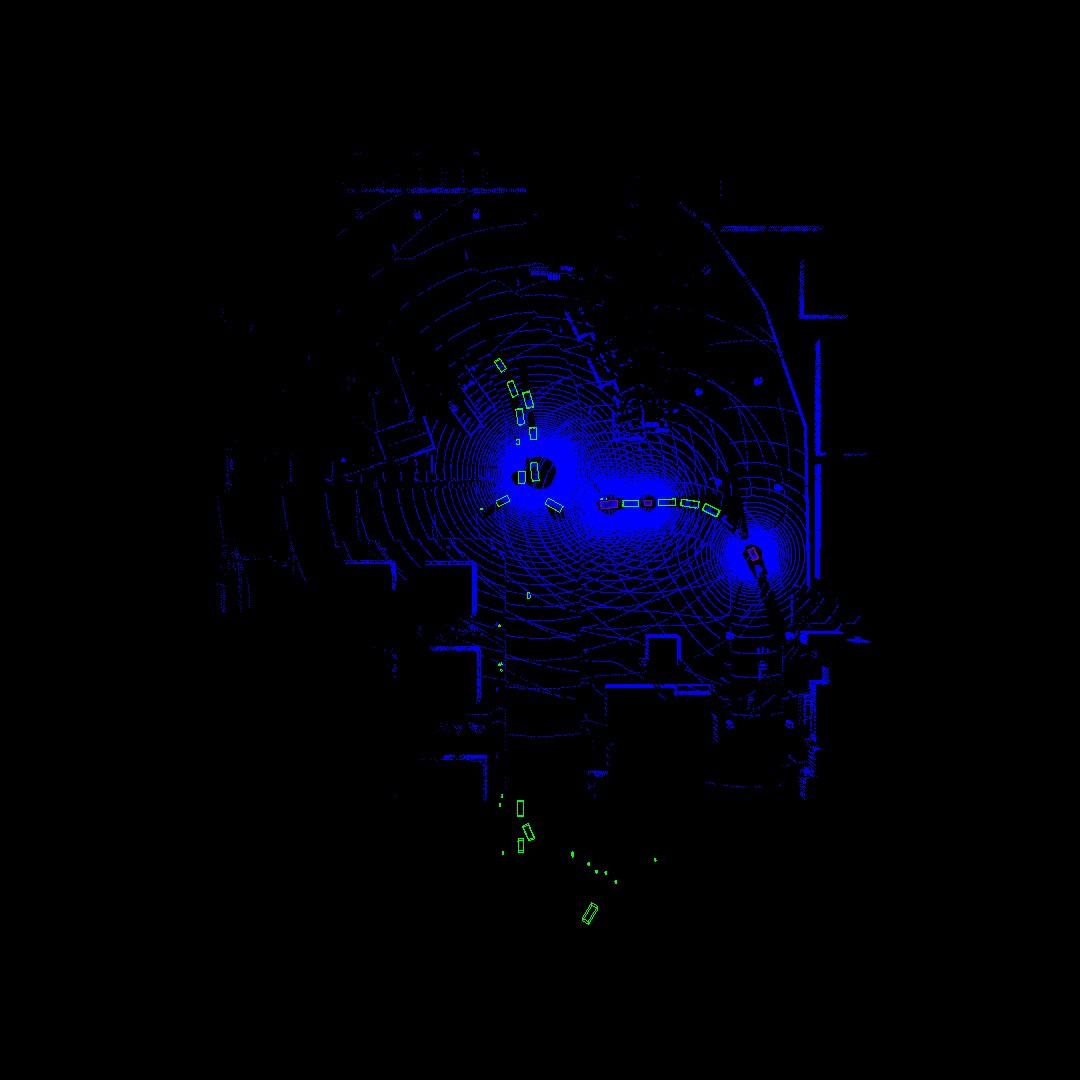} \\
            \includegraphics[width=\textwidth]{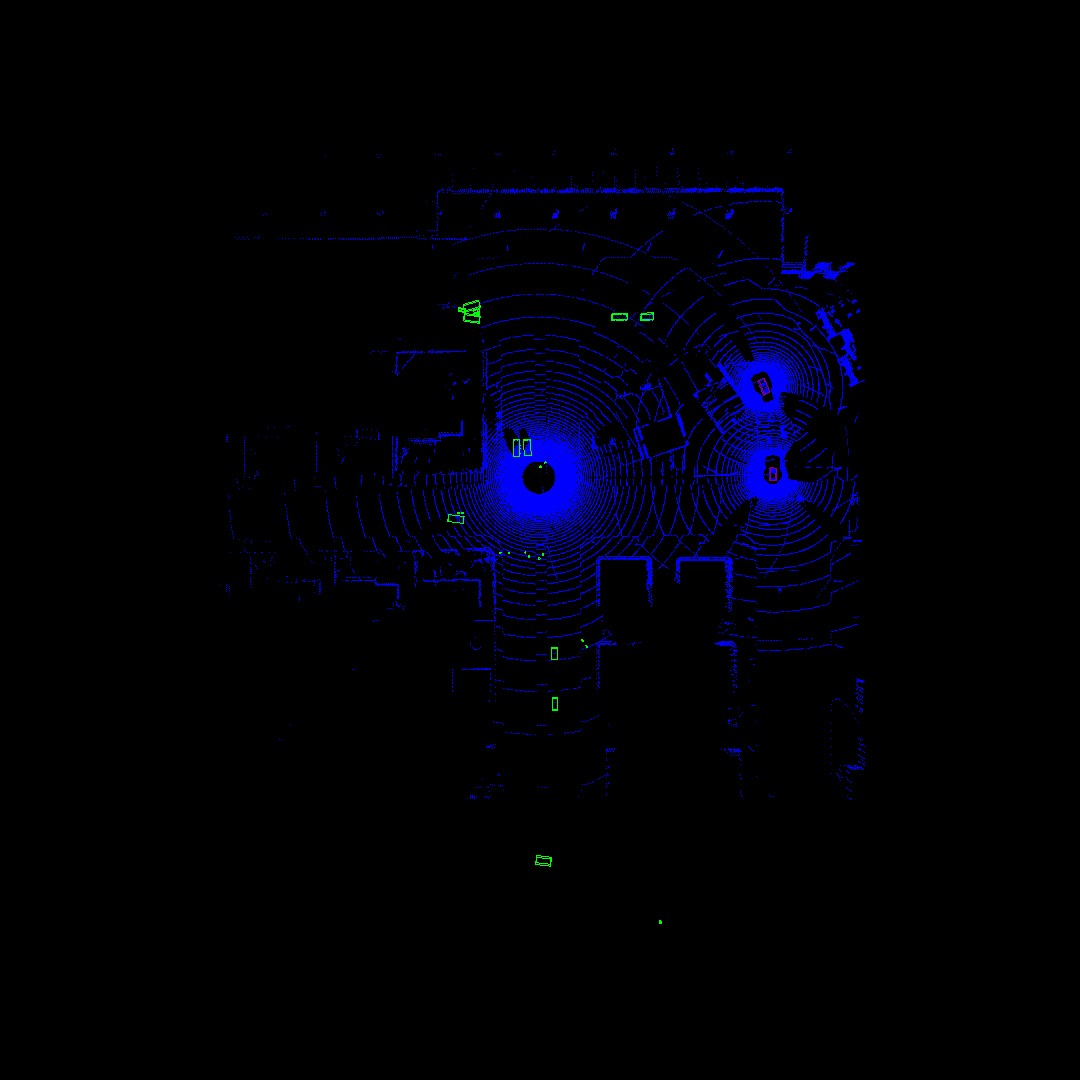} \\
            \includegraphics[width=\textwidth]{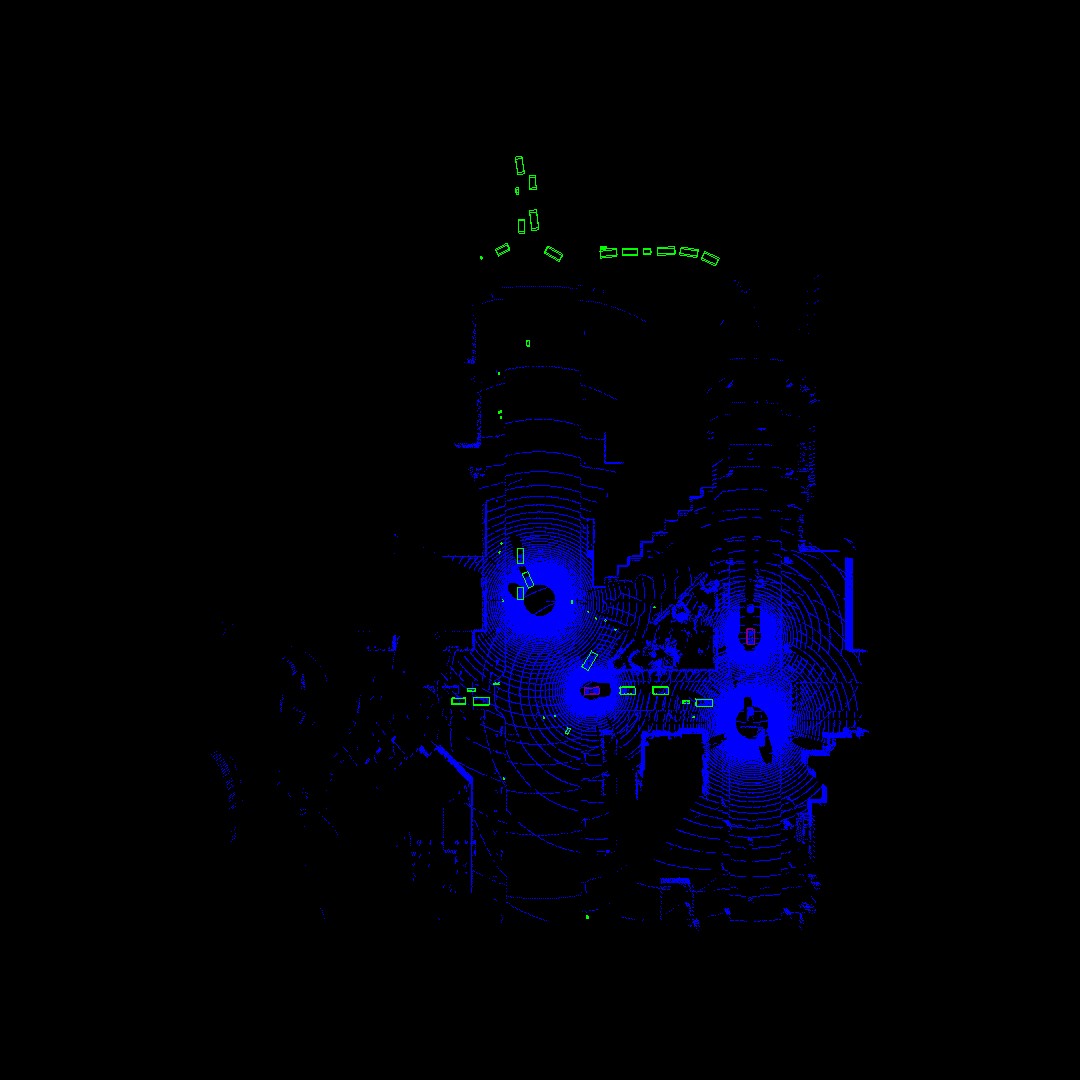} \\
            \includegraphics[width=\textwidth]{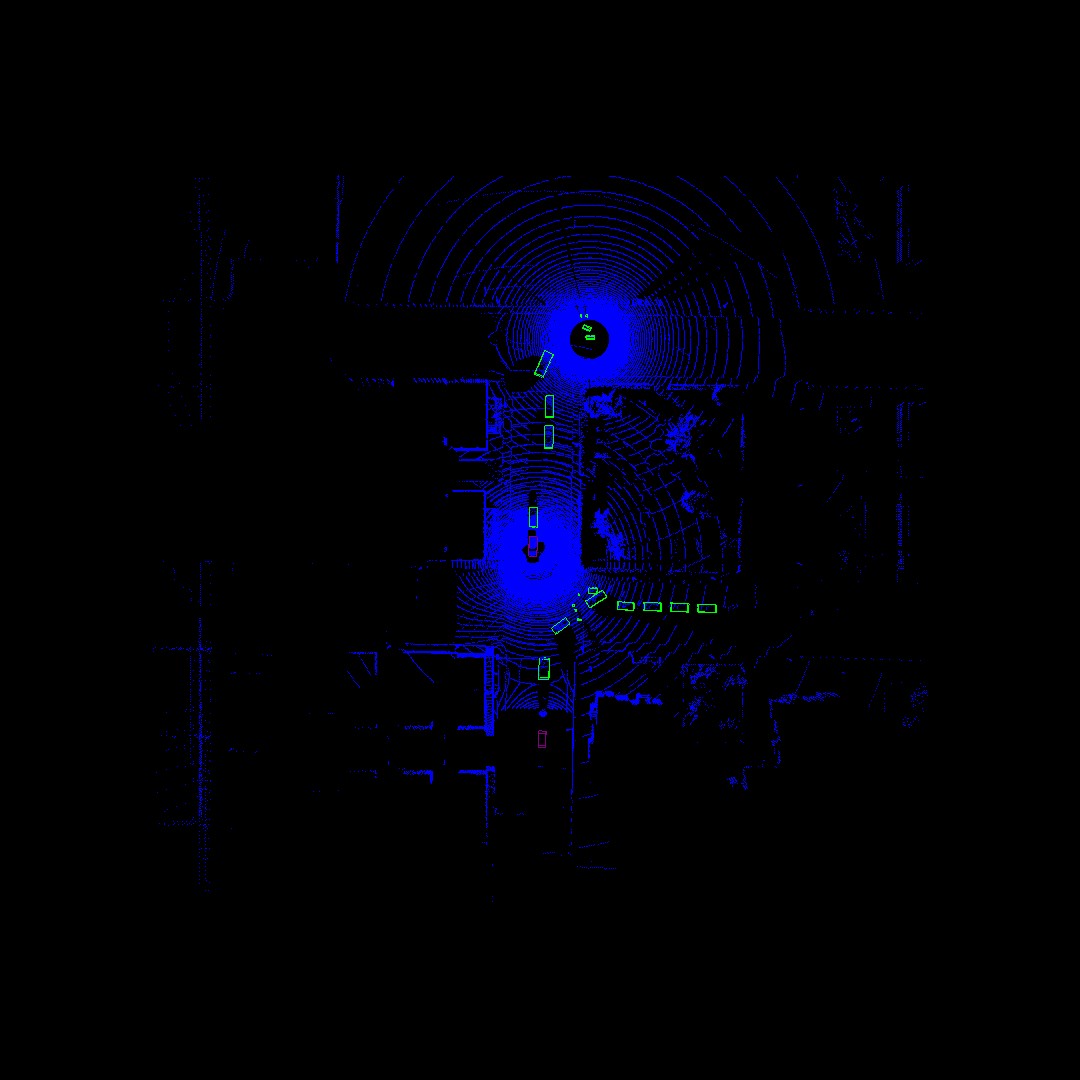} \\
            \caption{Town03}
        \end{minipage}
    \end{subfigure}
    \begin{subfigure}{0.30\textwidth}
        \begin{minipage}[c]{\textwidth}
            \centering
            \includegraphics[width=\textwidth]{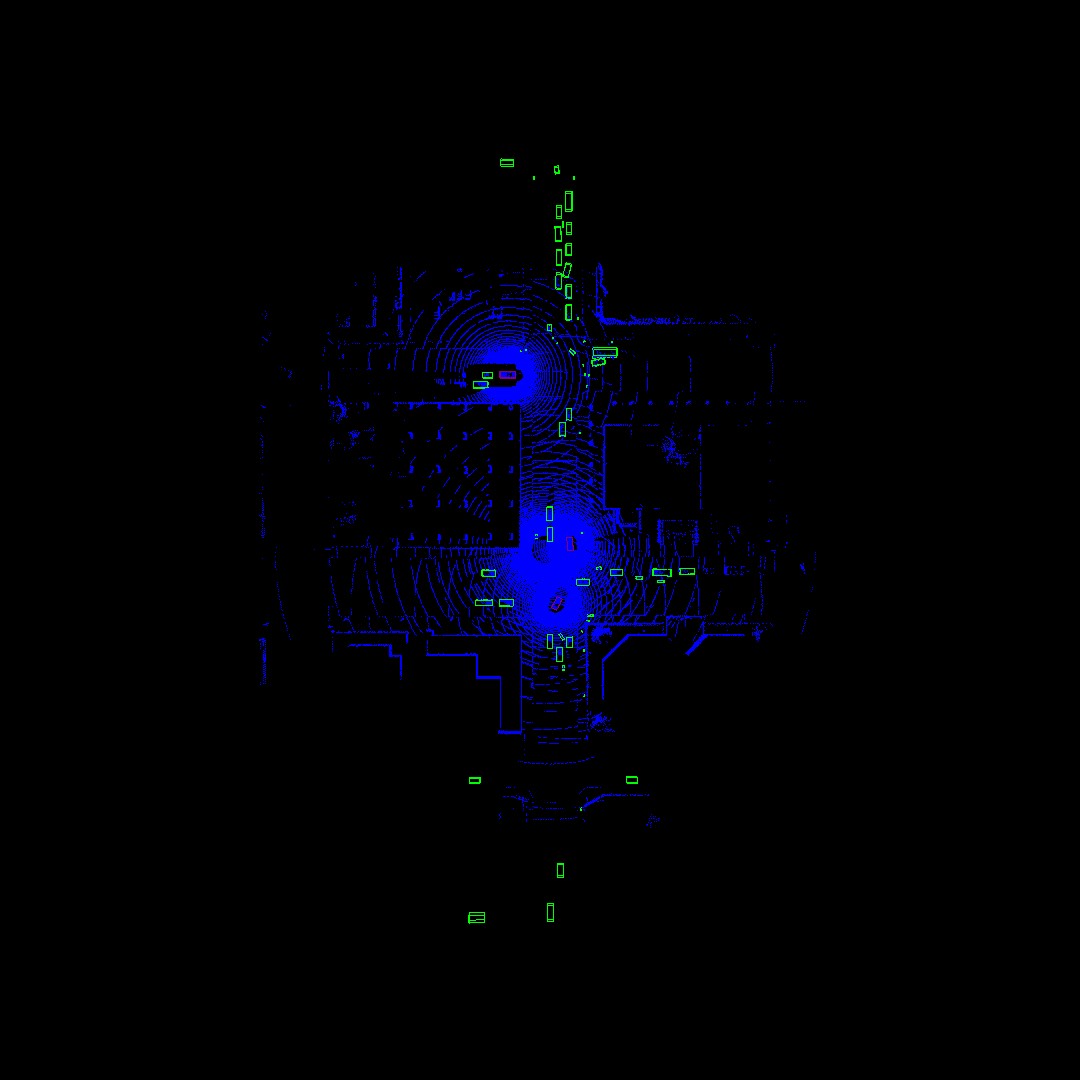} \\
            \includegraphics[width=\textwidth]{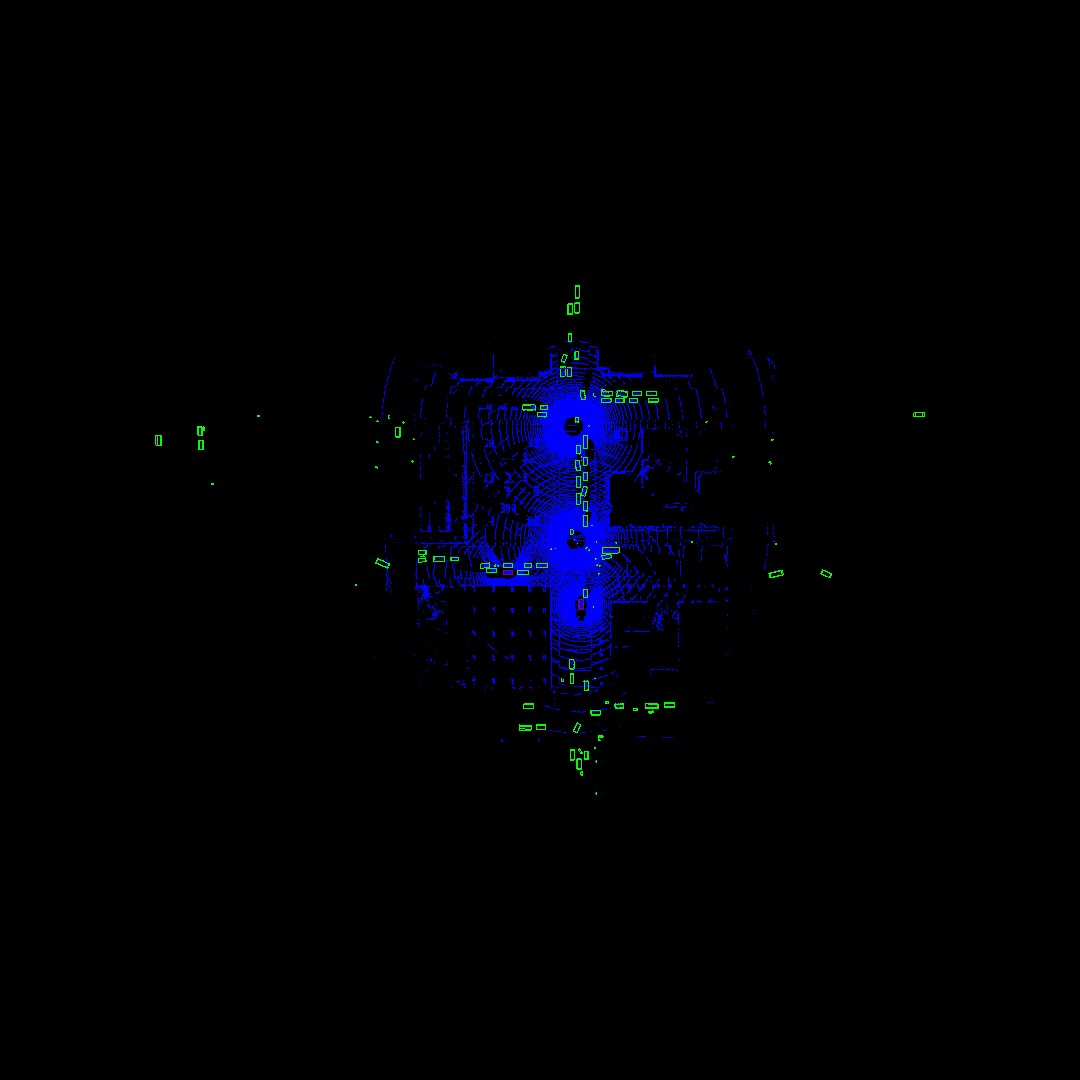} \\
            \includegraphics[width=\textwidth]{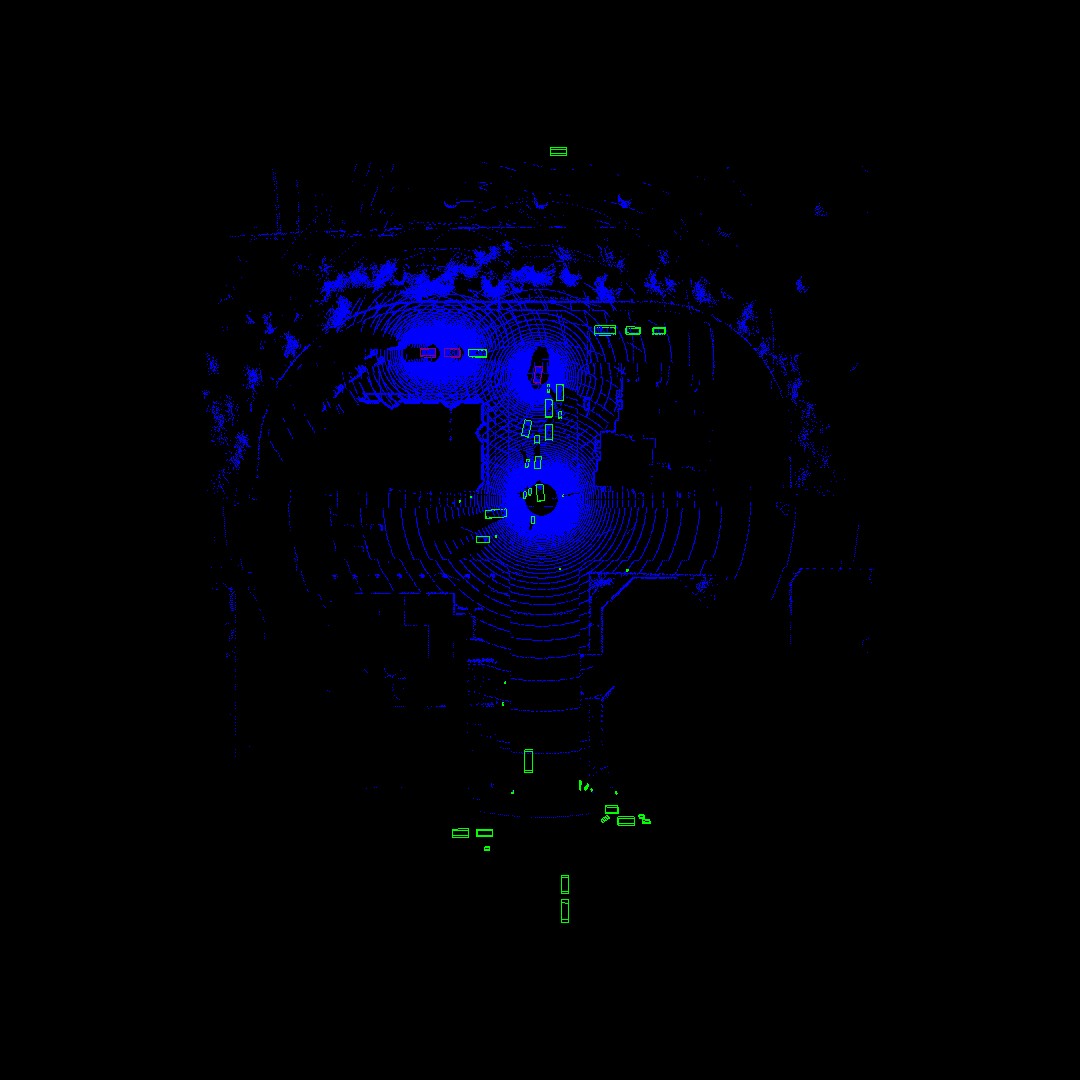} \\
            \includegraphics[width=\textwidth]{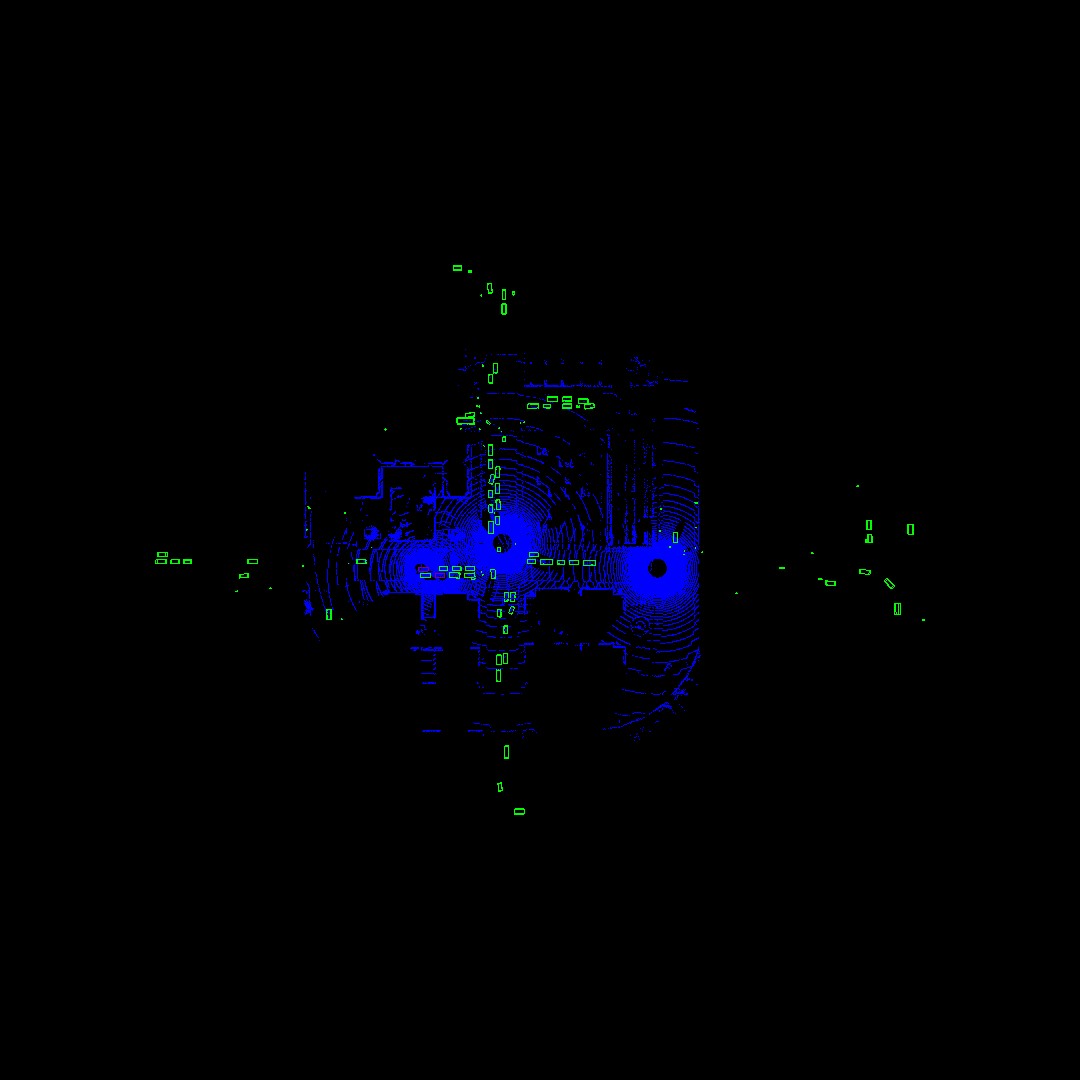} \\
            \caption{Town05}
        \end{minipage}
    \end{subfigure}
    \caption{Visualization examples of bird's eye view point cloud of Town01, Town03 and Town05 in Multi-V2X.}
    \label{fig:appendix-multiv2x-examples}
\end{figure*}

\begin{figure*}
    \centering
    \begin{subfigure}{0.30\textwidth}
        \begin{minipage}[c]{\textwidth}
            \centering
            \includegraphics[width=\textwidth]{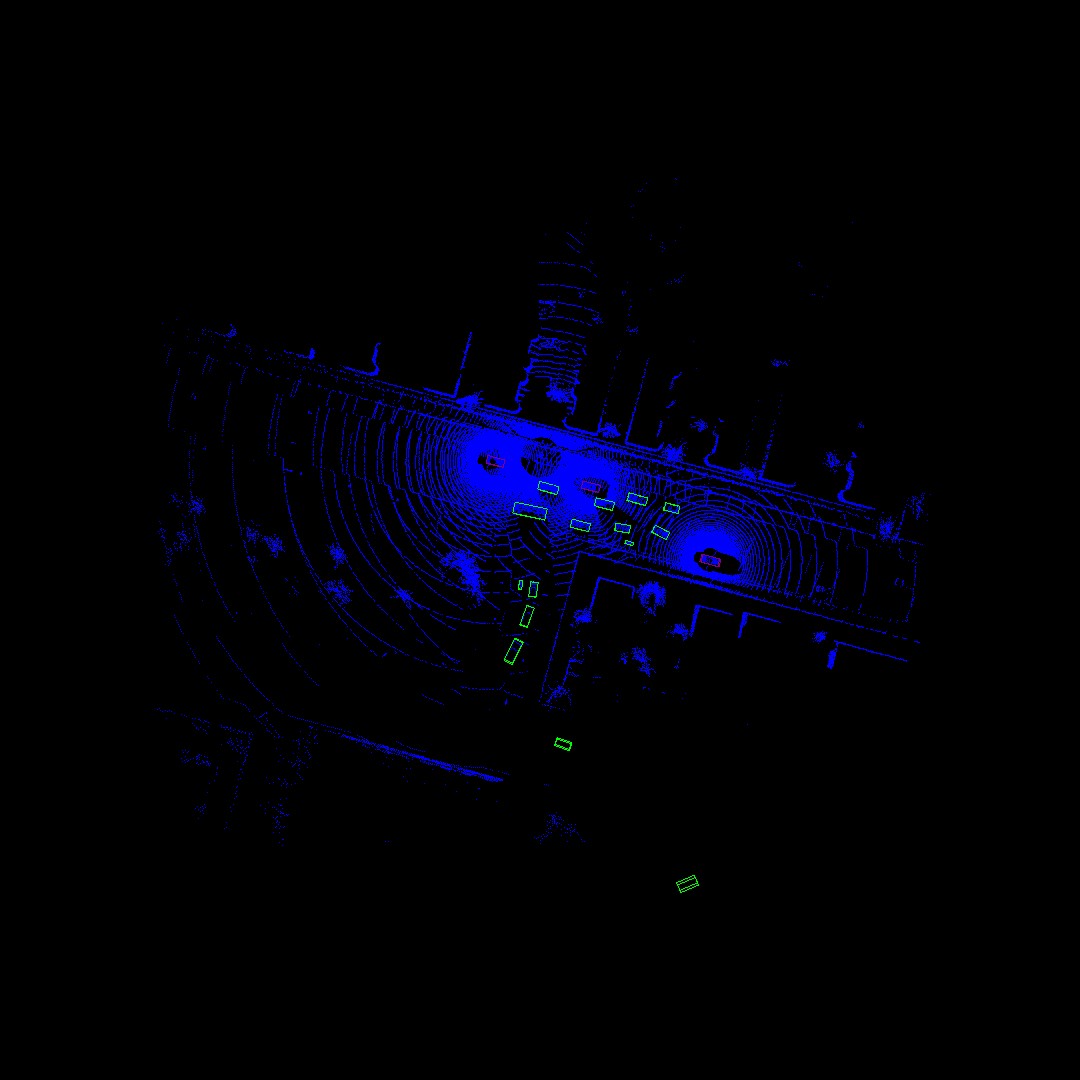} \\
            \includegraphics[width=\textwidth]{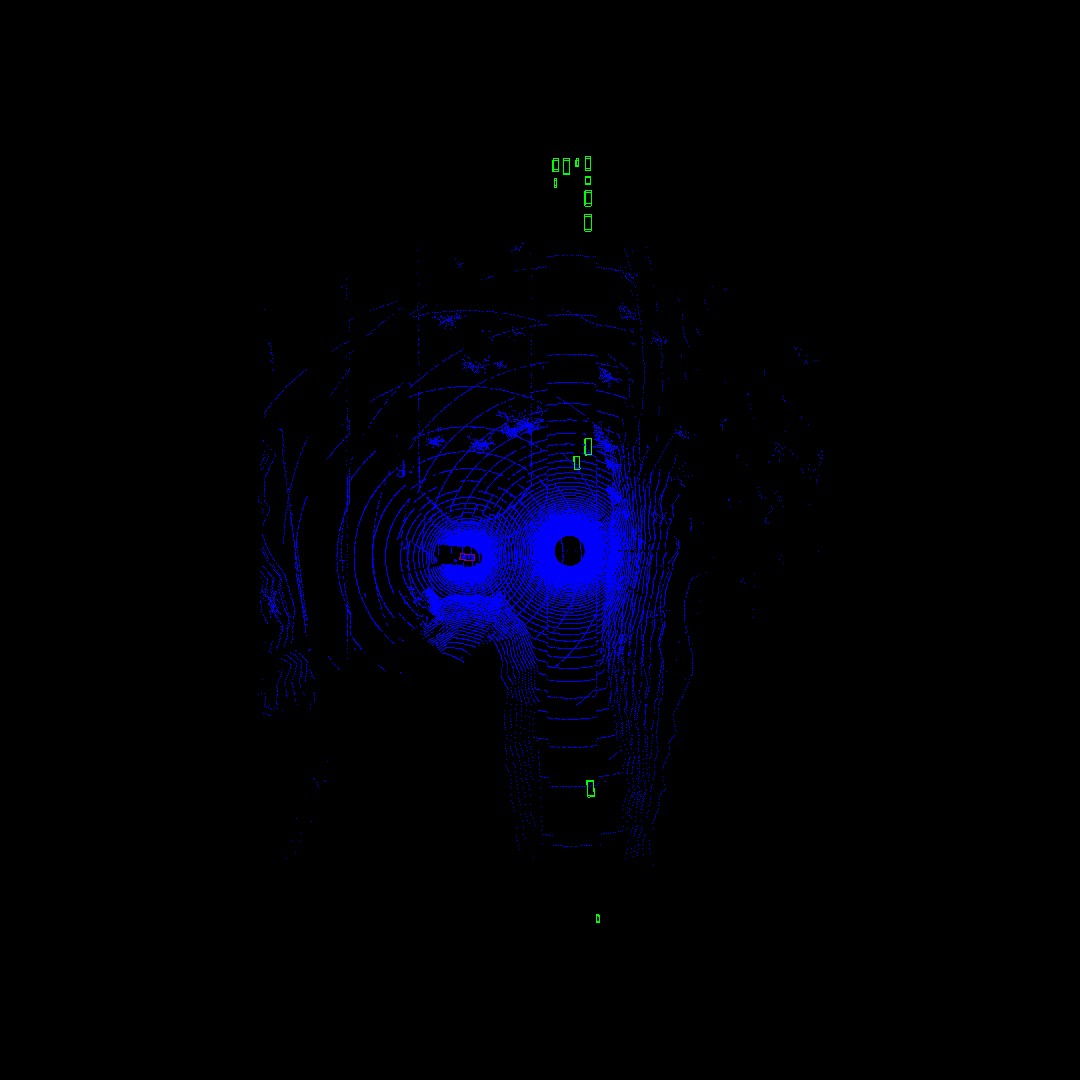} \\
            \includegraphics[width=\textwidth]{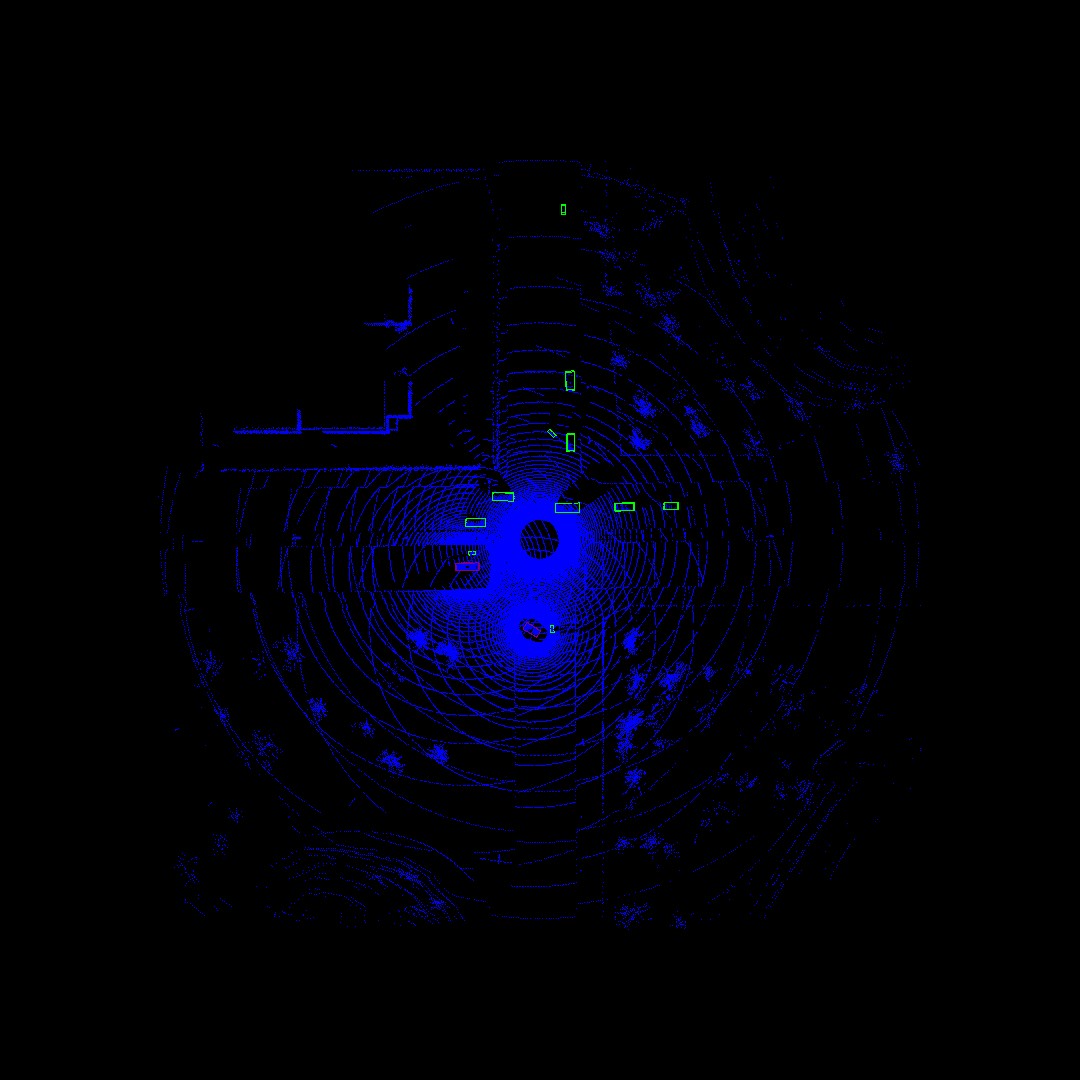} \\
            \includegraphics[width=\textwidth]{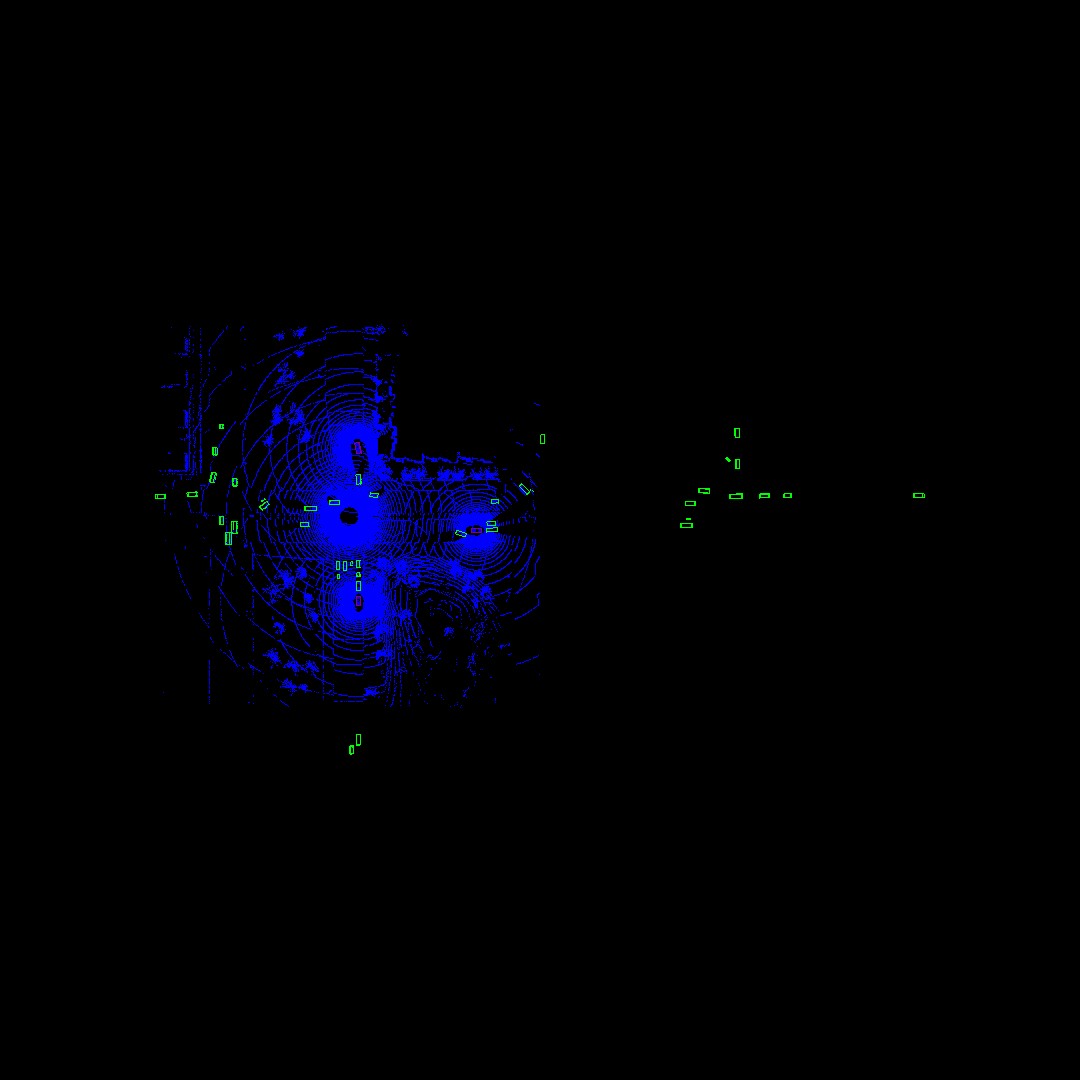} \\
            \caption{Town06}
        \end{minipage}
    \end{subfigure}
    \begin{subfigure}{0.30\textwidth}
        \begin{minipage}[c]{\textwidth}
            \centering
            \includegraphics[width=\textwidth]{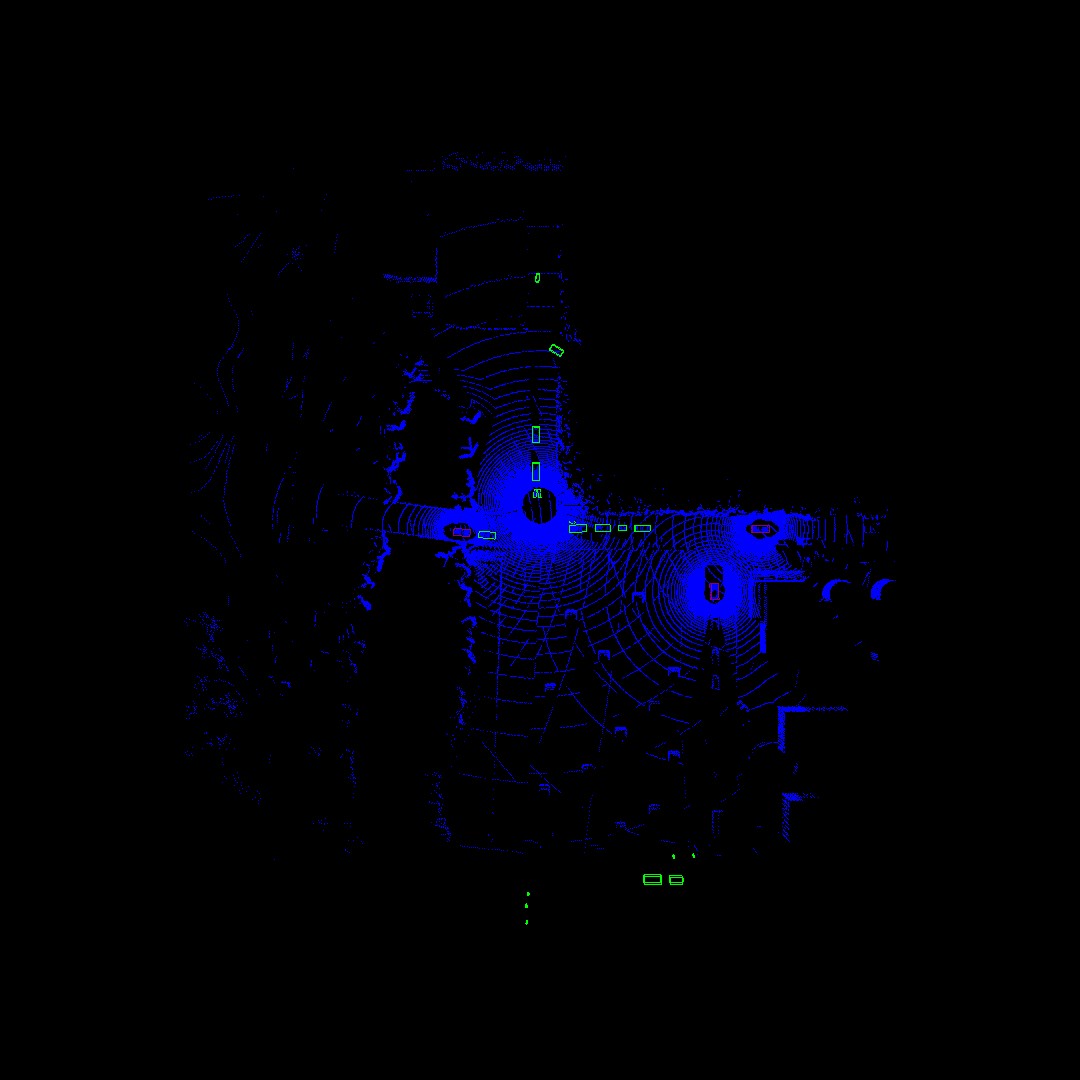} \\
            \includegraphics[width=\textwidth]{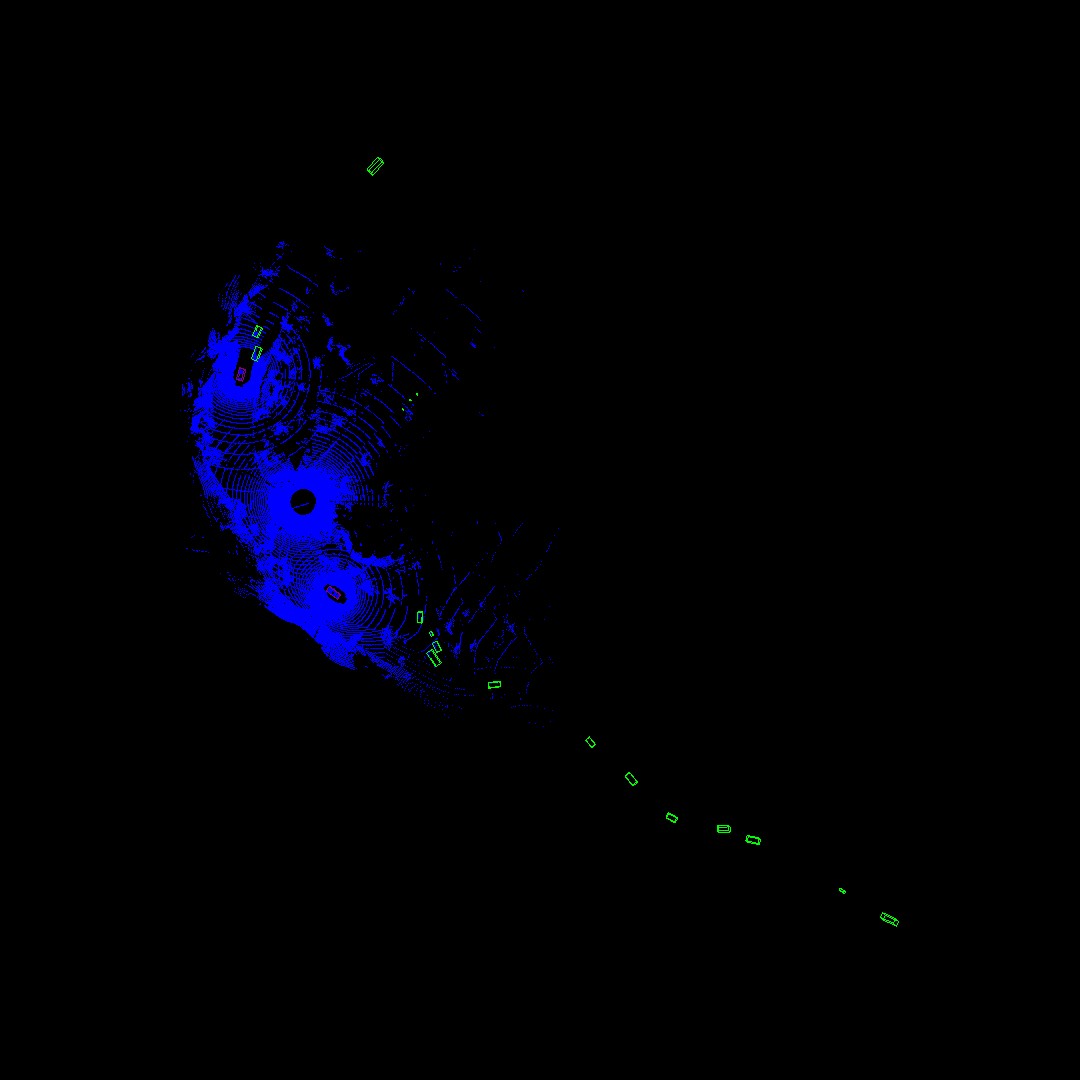} \\
            \includegraphics[width=\textwidth]{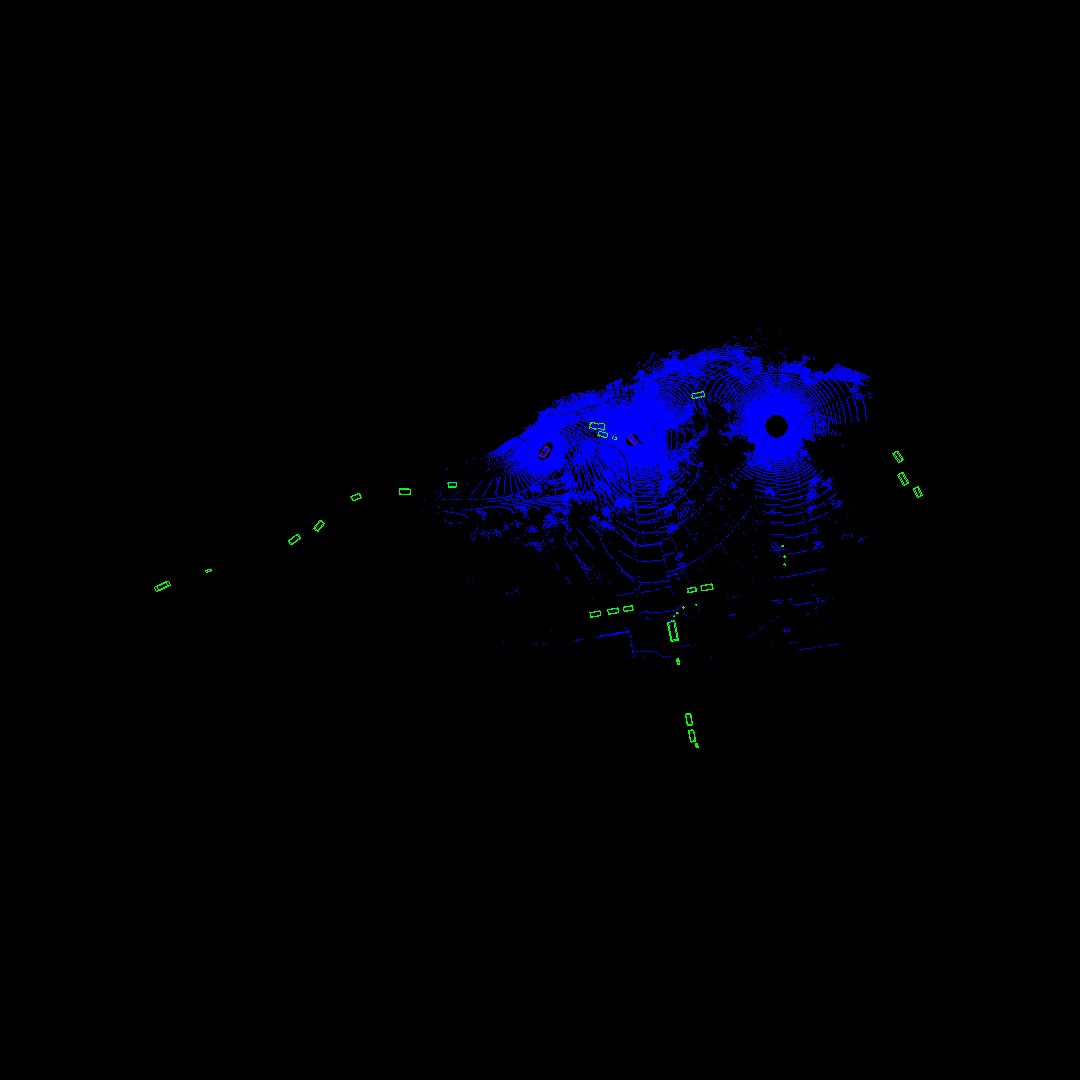} \\
            \includegraphics[width=\textwidth]{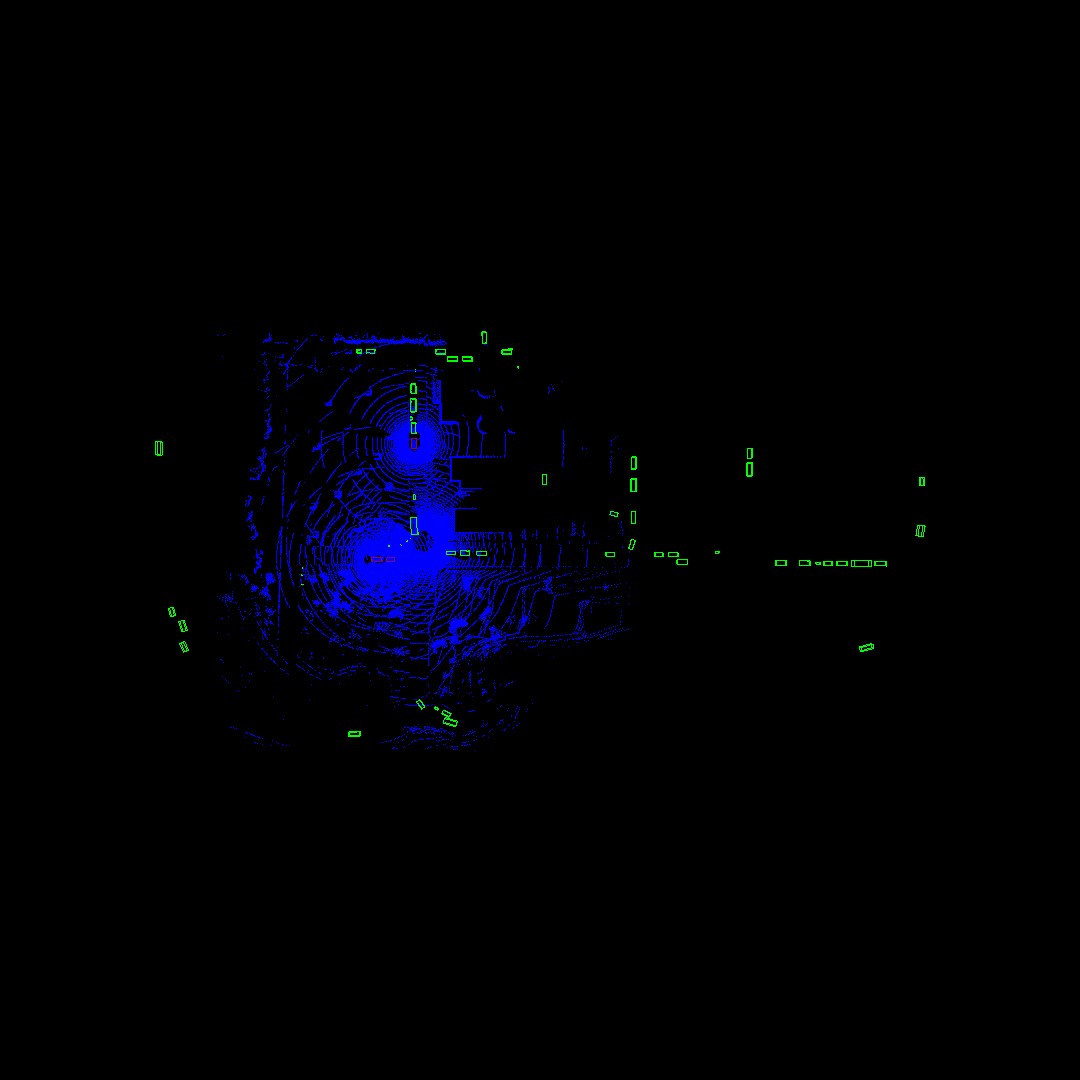} \\
            \caption{Town07}
        \end{minipage}
    \end{subfigure}
    \begin{subfigure}{0.30\textwidth}
        \begin{minipage}[c]{\textwidth}
            \centering
            \includegraphics[width=\textwidth]{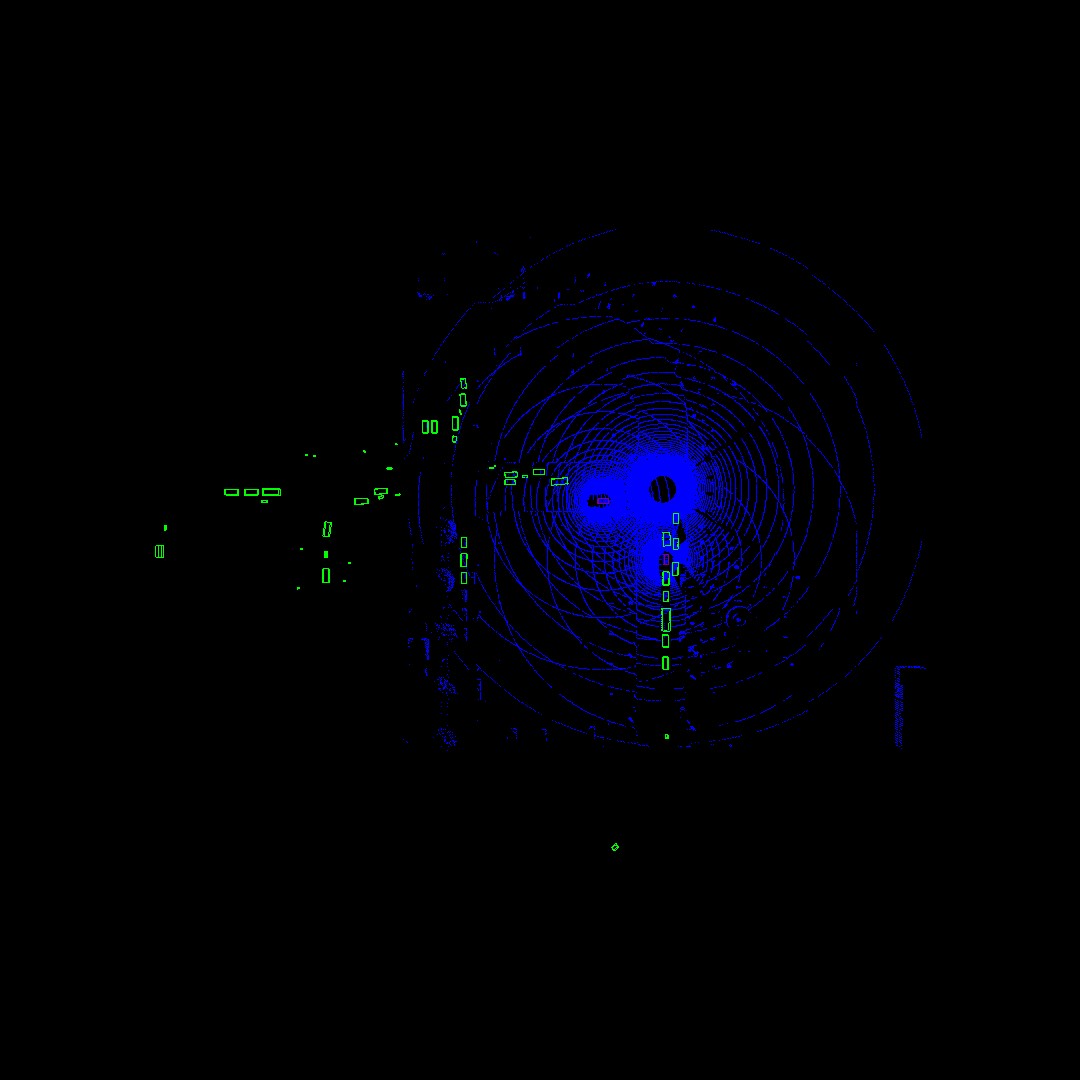} \\
            \includegraphics[width=\textwidth]{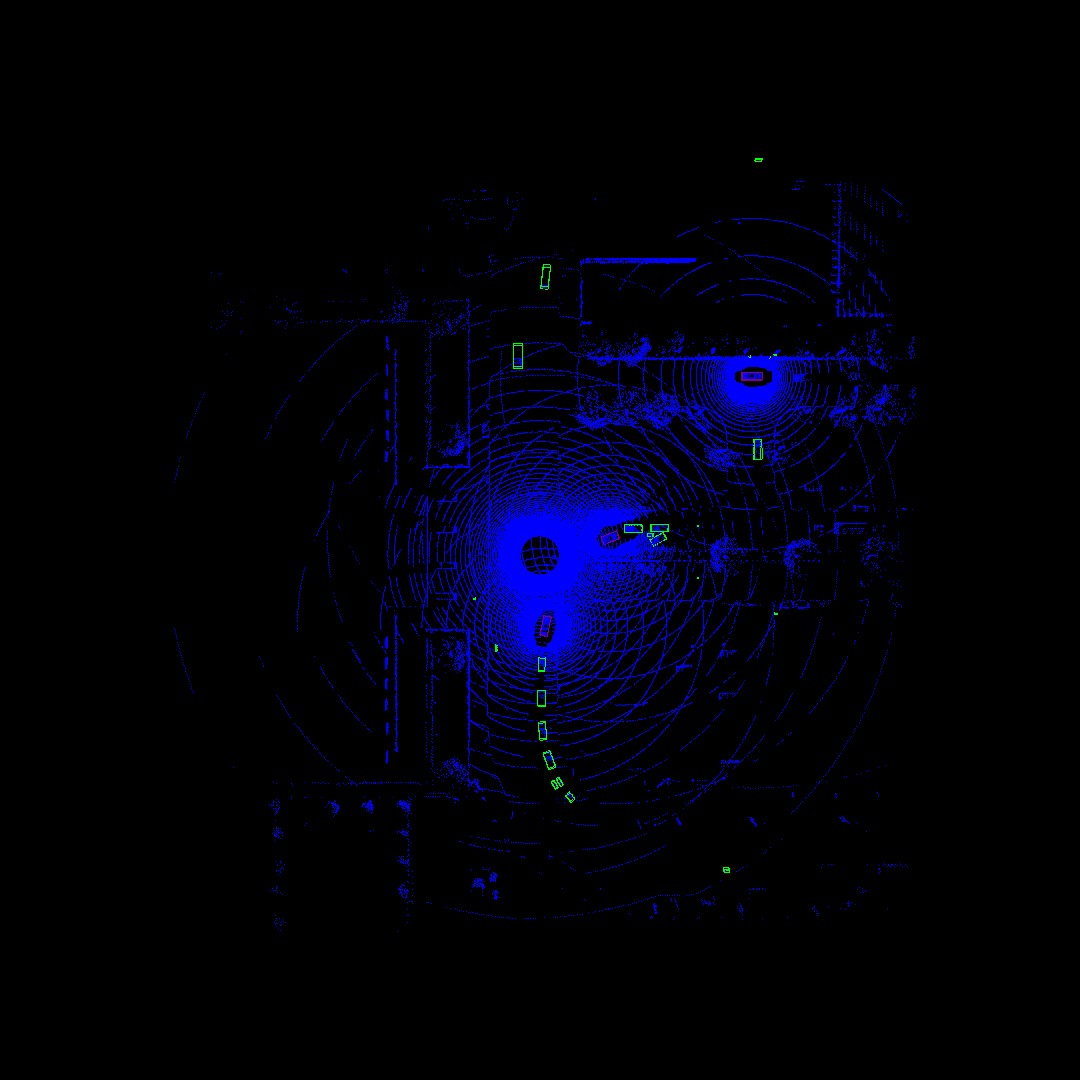} \\
            \includegraphics[width=\textwidth]{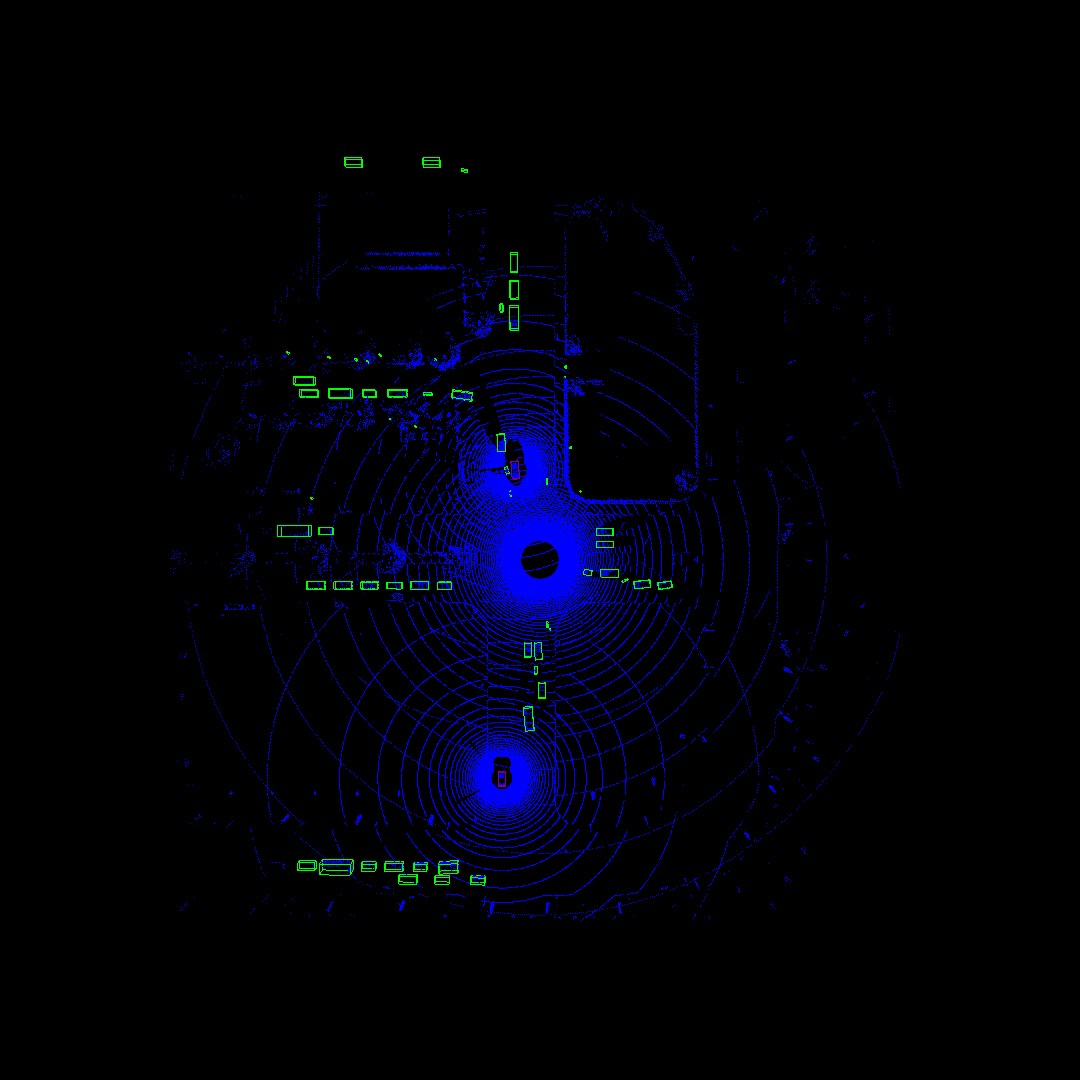} \\
            \includegraphics[width=\textwidth]{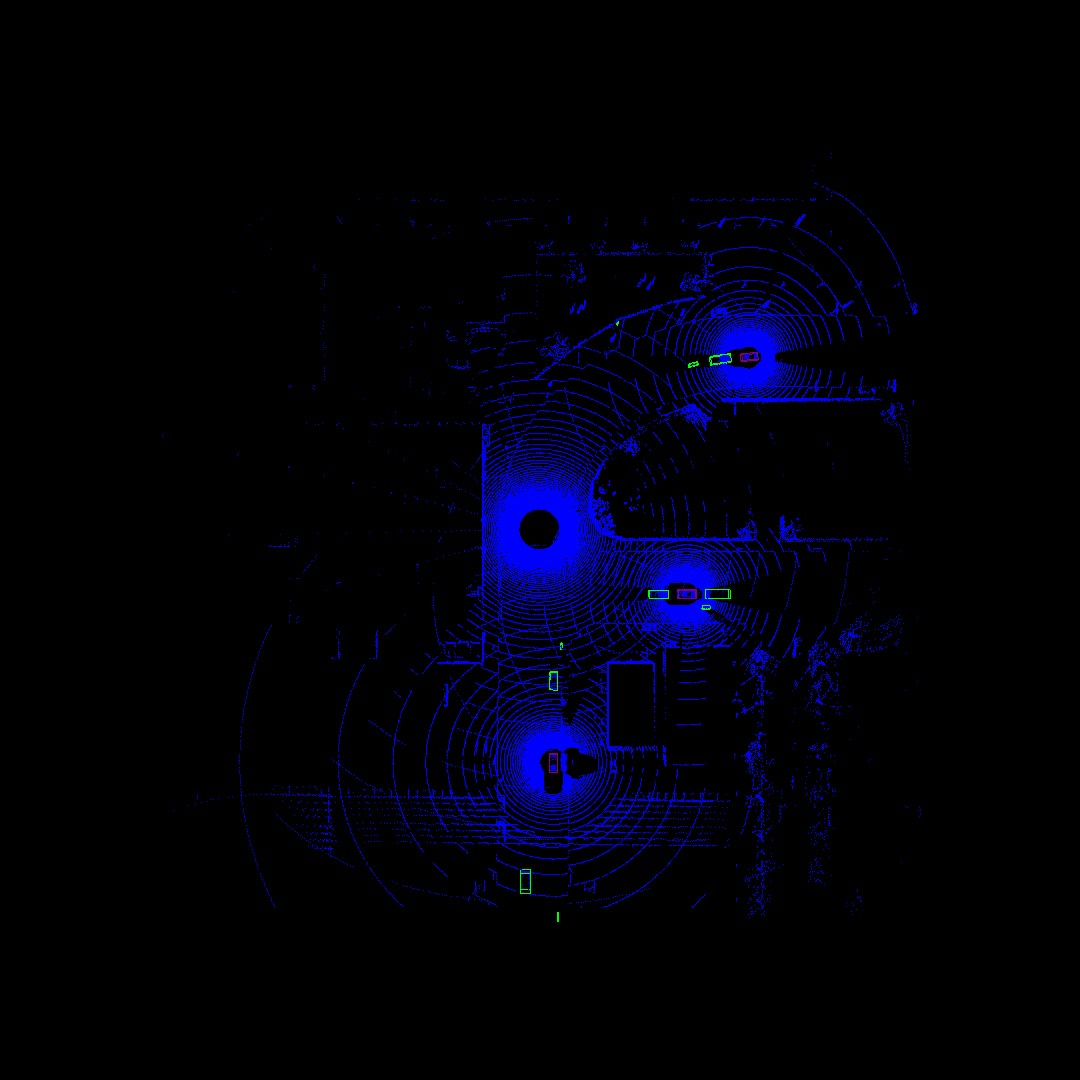} \\
            \caption{Town10HD}
        \end{minipage}
    \end{subfigure}
    \caption{Visualization examples of bird's eye view point cloud of Town06, Town07 and Town10HD in Multi-V2X.}
    \label{fig:appendix-multiv2x-examples-2}
\end{figure*}
 \fi

\end{document}